\documentclass{article} 
\usepackage{iclr2024_conference,times}

\usepackage{url}
\usepackage{booktabs} 
\usepackage{algorithm}
\usepackage{algpseudocode}
\usepackage{algorithmicx}
\algtext*{EndWhile}
\algtext*{EndFor}
\usepackage{graphicx}
\usepackage{wrapfig}
\usepackage{caption}
\usepackage{subcaption}
\usepackage{multicol}
\usepackage{multirow}
\usepackage{paralist}
\usepackage[utf8]{inputenc}
\usepackage{minitoc}
\usepackage{xspace}
\usepackage{xcolor}
\usepackage[breakable]{tcolorbox}
\usepackage{wrapfig}

\newcommand{\method}{RoboTool\xspace} 
\definecolor{citecolor}{RGB}{86,151,137}
\usepackage[pagebackref=true,breaklinks=true,colorlinks,citecolor=gray, linkcolor=citecolor]{hyperref} 
\hypersetup{
    citecolor = {citecolor}
}
\usepackage{url}
\usepackage{enumitem}

\title{Creative Robot Tool Use with Large Language Models}

\author{Mengdi Xu$^1$$^*$$^\diamond$, Peide Huang$^1$$^*$, Wenhao Yu$^2$$^*$, Shiqi Liu$^1$, Xilun Zhang$^1$, Yaru Niu$^1$,\\
  \textbf{Tingnan Zhang$^2$, Fei Xia$^2$, Jie Tan$^2$, Ding Zhao$^1$}\\
  \quad \\
  $^1$Carnegie Mellon University, $^2$Google DeepMind \\
  \texttt{\{mengdixu, peideh, shiqiliu, xilunz, yarun, dingzhao\}@andrew.cmu.edu} \\
  \texttt{\{magicmelon, tingnan, xiafei, jietan\}@google.com }
}

\newcommand\blfootnote[1]{%
  \begingroup
  \renewcommand\thefootnote{}\footnote{#1}%
  \addtocounter{footnote}{-1}%
  \endgroup
}

\iclrfinalcopy 
\begin{document}

\maketitle

\blfootnote{$^\ast$ Equal Contribution. $^\diamond$ Work partially done at Google DeepMind. }


\begin{abstract}
Tool use is a hallmark of advanced intelligence, exemplified in both animal behavior and robotic capabilities. This paper investigates the feasibility of imbuing robots with the ability to creatively use tools in tasks that involve implicit physical constraints and long-term planning. Leveraging Large Language Models (LLMs), we develop \method, a system that accepts natural language instructions and outputs executable code for controlling robots in both simulated and real-world environments. \method incorporates four pivotal components: (i) an ``Analyzer" that interprets natural language to discern key task-related concepts, (ii) a ``Planner" that generates comprehensive strategies based on the language input and key concepts, (iii) a ``Calculator" that computes parameters for each skill, and (iv) a ``Coder" that translates these plans into executable Python code. Our results show that \method can not only comprehend explicit or implicit physical constraints and environmental factors but also demonstrate creative tool use. Unlike traditional Task and Motion Planning (TAMP) methods that rely on explicit optimization, our LLM-based system offers a more flexible, efficient, and user-friendly solution for complex robotics tasks. Through extensive experiments, we validate that \method is proficient in handling tasks that would otherwise be infeasible without the creative use of tools, thereby expanding the capabilities of robotic systems. Demos are available on our project page: \url{https://creative-robotool.github.io/}.
\end{abstract}

\section{Introduction}
\label{sec:intro}

Tool use is an important hallmark of advanced intelligence. Some animals can use tools to achieve goals that are infeasible without tools. For example, Koehler's apes stacked crates together to reach a high-hanging banana bunch~\citep{kohler2018mentality}, and the crab-eating macaques used stone tools to open nuts and bivalves~\citep{gumert2009physical}. Beyond using tools for their intended purpose and following established procedures, using tools in creative and unconventional ways provides more flexible solutions, albeit presents far more challenges in cognitive ability. In robotics, creative tool use \citep{fitzgerald2021modeling} is also a crucial yet very demanding capability because it necessitates the all-around ability to predict the outcome of an action, reason what tools to use, and plan how to use them. In this work, we want to explore the question, \textit{can we enable such creative tool-use capability in robots?} We identify that creative robot tool use solves a complex long-horizon planning task with constraints related to environment and robot capacity. 
For example, ``grasping a milk carton" while the milk carton's location is out of the robotic arm's workspace or ``walking to the other sofa" while there exists a gap in the way that exceeds the quadrupedal robot's walking capability.

Task and motion planning (TAMP) is a common framework for solving such long-horizon planning tasks. It combines low-level continuous motion planning and high-level discrete task planning to solve complex planning tasks that are difficult to address by any of these domains alone. Existing literature can handle tool use in a static environment with optimization-based approaches such as logic-geometric programming~\citep{toussaint2018differentiable}. However, this optimization approach generally requires a long computation time for tasks with many objects and task planning steps due to the increasing search space~\citep{garrett2021integrated}. In addition, classical TAMP methods are limited to the family of tasks that can be expressed in formal logic and symbolic representation, making them not user-friendly for non-experts~\citep{chen2023autotamp, lin2023text2motion}.

Recently, large language models (LLMs) have been shown to encode vast knowledge beneficial to robotics tasks in reasoning, planning, and acting~\citep{brohan2023rt, huang2023voxposer, yu2023language}. TAMP methods with LLMs are able to bypass the computation burden of the explicit optimization process in classical TAMP. Prior works show that LLMs can adeptly dissect tasks given either clear or ambiguous language descriptions and instructions. Robots powered by LLMs also demonstrate notable compositional generalization in TAMP~\citep{huang2022language, ahn2022can}. However, it is still unclear how to use LLMs to solve more complex tasks that require reasoning with implicit constraints imposed by the robot's embodiment and its surrounding physical world.

\begin{figure}[t]
\vspace{-1.0cm}
\begin{subfigure}[b]{0.385\textwidth}
    \includegraphics[width=1.0\linewidth]{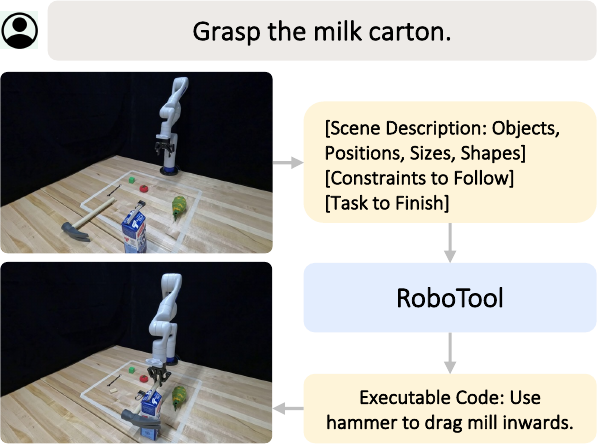}
    \caption{RoboTool Overview}
\end{subfigure}
\hfill
\begin{subfigure}[b]{0.6\textwidth}
    \includegraphics[width=1.0\linewidth]{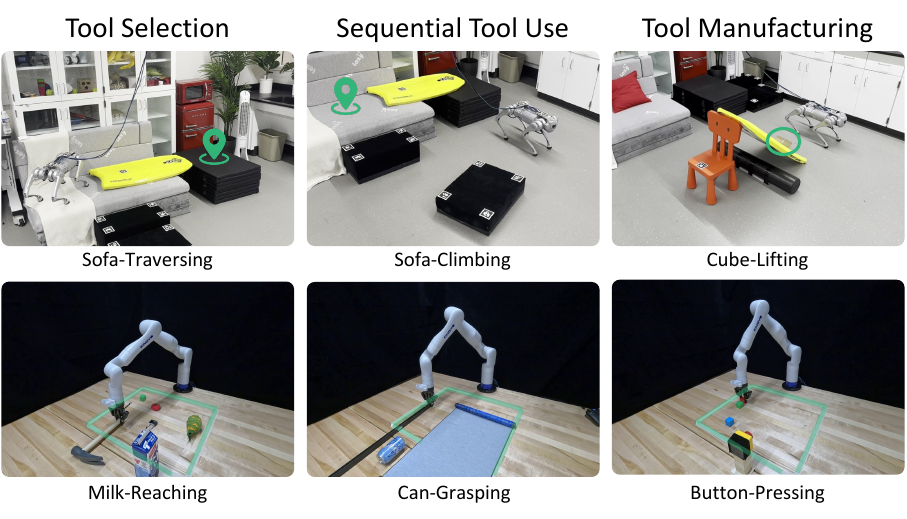}
    \vspace{-0.7cm}
    \caption{Creative Tool Use Benchmark}
    \label{fig:overview_benchmark}
\end{subfigure}
\vspace{-0.1in}
\caption{\small (a) Creative robot tool use with Large Language Models (\method). \method takes natural language descriptions as input, including the scene descriptions, environment- and embodiment-related constraints, and tasks. (b) We design a creative tool-use benchmark based on a quadrupedal robot and a robotic arm, including 6 challenging tasks that symbols three types of creative tool-use behaviors.
}
\label{fig:overview}
\vspace{-0.1in}
\end{figure}

In this work, we are interested in solving language-instructed long-horizon robotics tasks with implicitly activated physical constraints (Fig.~\ref{fig:overview}).
By providing LLMs with adequate numerical semantic information in natural language, we observe that LLMs can identify the activated constraints induced by the spatial layout of objects in the scene and the robot's embodiment limits, suggesting that LLMs may maintain knowledge and reasoning capability about the 3D physical world.
For example, in the Sofa-Traversing example, the LLM can identify that the key concept affecting the plan's feasibility is the gap width between two sofas, although there is no prior information about the existence of the ``gap" concept in the provided language instruction.
Furthermore, our comprehensive tests reveal that LLMs are not only adept at employing tools to transform otherwise unfeasible tasks into feasible ones but also display creativity in using tools beyond their conventional functions, based on their material, shape, and geometric features.
Again, in the Sofa-Traversing example, the LLM could use the surfboard next to the quadrupedal robot as a bridge to walk across the gap.

To solve the aforementioned problem, we introduce \method, a creative robot tool user built on LLMs, which uses tools beyond their standard affordances.
\method accepts natural language instructions comprising textual and numerical information about the environment, robot embodiments, and constraints to follow. 
\method produces code that invokes robot's parameterized low-level skills to control both simulated and physical robots.  
\method consists of four central components, with each handling one functionality, as depicted in Fig.~\ref{fig:method}: (i) \textit{Analyzer}, which processes the natural language input to identify key concepts that could impact the task's feasibility, (ii) \textit{Planner}, which receives both the original language input and the identified key concepts to formulate a comprehensive strategy for completing the task, (iii) \textit{Calculator}, which is responsible for determining the parameters, such as the target positions required for each parameterized skill, and (iv) \textit{Coder}, which converts the comprehensive plan and parameters into executable code. All of these components are constructed using GPT-4.

Our key contributions are in three folds:
\begin{itemize}[leftmargin=0.7cm]
    \vspace{-0.05in}
    \item We introduce \method, a creative robot tool user based on pretrained LLMs, that can solve long-horizon hybrid discrete-continuous planning problems with environment- and embodiment-related constraints in a zero-shot manner.
    \vspace{-0.01in}
    \item We provide an evaluation benchmark (Fig.~\ref{fig:overview_benchmark}) to test various aspects of creative tool-use capability, including tool selection, sequential tool use, and tool manufacturing, across two widely used embodiments: the robotic arm and the quadrupedal robot. 
    \vspace{-0.01in}
    \item Simulation and real-world experiments demonstrate that \method solves tasks unachievable without creative tool use and outperforms baselines by a large margin in terms of success rates.
\end{itemize}

\section{Related Works}
\label{sec:related_work}
\vspace{-0.1in}

\textbf{Language Models for Task and Motion Planning (TAMP).} 
TAMP~\citep{garrett2021integrated} has been integrated with LLMs for building intelligent robots. Most of the literature built upon hierarchical planning~\citep{garrett2020pddlstream, garrett2021integrated, kaelbling2011hierarchical}, where LLMs only provide a high-level plan that invokes human-engineered control primitives or motion planners~\citep{ahn2022can, huang2022language, huang2022inner, ren2023robots, chen2023autotamp, ding2023task, silver2023generalized, liu2023llm+, xie2023translating}. 
In this work, we follow the hierarchical planning setting and aim to develop an LLM-based planner to solve tasks with constraints that require creative tool-use behaviors.
One challenge of using LLMs as a planner is to ground it with real-world interactions. SayCan~\citep{ahn2022can}, Grounded Decoding~\citep{huang2023grounded} and Text2Motion~\citep{lin2023text2motion} grounded an LLM planner with a real-world affordance function (either skill-, object- or environment-related) to propose feasible and appropriate plans. Developing these affordance functions requires extra training from massive offline data or domain-specific knowledge. In contrast, we rely entirely on LLM's capability of deriving the affordance from the language input and do not require separate pretrained affordance functions.

Existing works integrating LLM and TAMP output the plan either in the format of natural language, PDDL language, or code scripts. With the focus on robotics applications, one natural interface to call the low-level skills is code generated by LLMs.
Code-as-Policies~\citep{liang2023code} and ProgPrompt~\citep{singh2023progprompt} showed that LLMs exhibit spatial-geometric reasoning and assign precise values to ambiguous descriptions as well as writing snippets of logical Python code.
In the multi-modal robotics, Instruct2Act~\citep{huang2023instruct2act} and VoxPoser~\cite{huang2023voxposer} also generate code to incorporate perception, planning, and action. Following these works, we propose to use a standalone LLM module to generate codes.

\textbf{Robot Tool Use.}
Tool use enables robots to solve problems that they were unable to without tools. There is a long history of interest in robotics literature that focuses on manipulating tools to finish designated tasks, such as furniture polishing~\citep{nagata2001furniture}, nut fastening~\citep{pfeiffer2017nut}, playing table tennis~\citep{muelling2010learning}, using chopsticks~\cite{ke2021grasping} etc. Despite these successful attempts, they focus on generating actions for specific tools and do not study the causal effect of tools and their interaction with other objects. Therefore, they cannot deal with novel objects or tasks and generally lack improvisational capability.
To interact with tools in a more intelligent way, \cite{sinapov2008detecting} and \cite{levihn2014using} conducted early attempts to study the effects of different tools and mechanisms. 
\cite{wicaksono2016relational} developed a system to learn a simple tool from a demonstration of another agent employing a similar tool by generating and testing hypotheses, represented by Horn clauses, about what tool features are important.
\cite{xie2019improvisation} and \cite{fang2020learning} trained deep neural networks from diverse self-supervised data of tool use to either predict the effects of tools or generate the action directly.

Related to our method, \cite{toussaint2018differentiable} formulated a Logic-Geometric Program to solve physical puzzles with sequential tool use, such as using a hook to get another longer hook to reach for a target ball. 
\cite{ren2023leveraging} utilized LLMs to transform task and tool features in text form as latent representations and concatenate them with vision input to achieve faster adaptation and generalization via meta-learning. 
More recently, RT2 \citep{brohan2023rt} performed multi-stage semantic reasoning to, for example, decide the rock could be used as an improvised hammer. 
Unlike most of the existing robotics literature, our method leverages LLM's massive prior knowledge about object affordance and impressive planning capability to propose creative solutions to different kinds of physical puzzles. These solutions require highly complex reasoning and planning capability.

\vspace{-0.05in}
\section{Methodology}
\label{sec:method}
\vspace{-0.1in}

\begin{figure}[t]
\vspace{-0.5cm}
\centerline{\includegraphics[width=\linewidth]{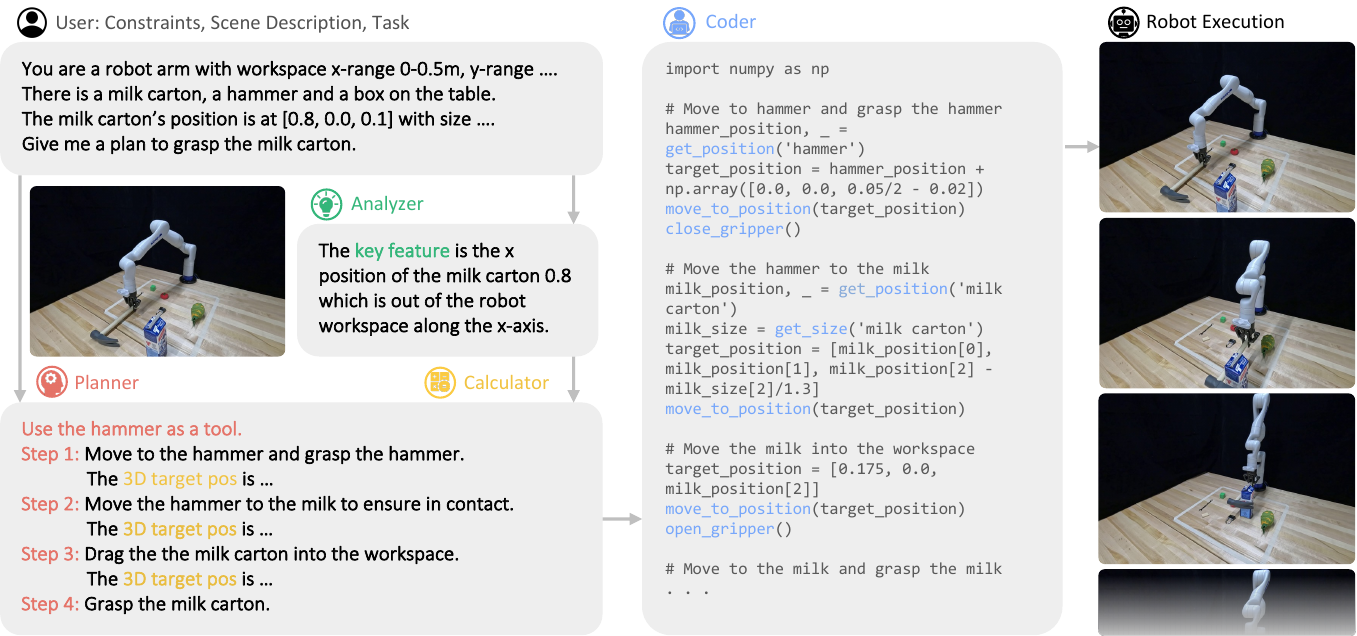}}
\vspace{-0.1in}
\caption{\small Overview of our proposed \method, which is a creative robot tool user consisting of four key components including \textit{Analyzer}, \textit{Planner}, \textit{Calculator} and \textit{Coder}.}
\vspace{-0.2in}
\label{fig:method}
\end{figure}

We are interested in enabling robots to solve complex long-horizon tasks with multiple environment- and embodiment-related constraints, that require robot's \textit{creative tool-use} capability to solve the tasks. In this section, we first posit our problem as a hybrid discrete-continuous planning problem in Sec.~\ref{sec:problem_formulation}. We then introduce our proposed method, \method, in Sec.~\ref{sec:method_details}, which is a creative robot tool user built on LLMs and can solve complex task planning problems in a zero-shot manner.

\vspace{-0.05in}
\subsection{Problem Formulation}
\label{sec:problem_formulation}
\vspace{-0.1in}
With a focus on robotic applications, we aim to solve a hybrid discrete-continuous planning problem with multiple constraints based on a natural language description. 
The provided description contains words and numerical values to depict the environments, tasks, and constraints.
Assuming the robot is equipped with a repertoire of parameterized skills, we seek a hierarchical solution to generate a plan composed of a sequence of provided skills and a sequence of parameters for the corresponding skills.
As code serves as a general interface to send commands to robots, our method will output executable robot policy code that sequentially calls skills with the parameters according to the plan, to complete the task.
Solving such problems typically involves interacting with different objects in the scene to satisfy the environment and embodiment constraints. 
This symbolizes \textit{the tool-use behavior}, which is a cornerstone of intelligence.

\textbf{Language Description as Input.} 
We define the environment layout space as $\mathcal{Q}$ and the initial environment configuration as $q_0 \in \mathcal{Q}$. 
The API $F_q$ helps parse $q_0$ into an environment language description $L_{\mathcal{Q}} = F_q(q_0)$, which includes the spatial layouts of objects, such as ``there is a hammer on the table" and ``the robot is on the ground," as well as each object's positions, sizes, and physical properties. 
We denote the constraint set as $\mathcal{C} = \mathcal{C_Q} \cup \mathcal{C_R}$, where $\mathcal{C_Q}$ represents the constraints related to environments, such as ``the scroll cannot be lifted", and $\mathcal{C_R}$ represents the constraints stemming from the robot's physical limitations, encompassing aspects like the robot's workspace boundary and the extent of its skills.
Let the robot embodiment space be $\mathcal{R}$. 
Each constraint $C\in\mathcal{C}$ can be activated based on different combination of $\mathcal{Q}$ and $\mathcal
R$.
The API $F_{\mathcal{C}}$ helps parse the constraints $\mathcal{C}$ into a constraint description $L_{\mathcal{C}} = F_{\mathcal{C}}(\mathcal{C})$. 
The user will provide the task $L_{\mathcal{T}}$ in natural language.
The concatenated language description $L = \{L_{\mathcal{T}}, L_{\mathcal{Q}}, L_{\mathcal{C}}\}$ serves as the query input to our method.

\textbf{Hierarchical Policies for Robot Tool Use.} We consider a Markov Decision Process $\mathcal{M}$ defined by a tuple
$(\mathcal{S}, \mathcal{A}, p, r, \rho_0)$, representing the state space, action space, transition dynamics, reward function, and initial state distribution, respectively. 
We use a two-level hierarchy consisting of a set of parameterized skills and a high-level controller.
For each robot embodiment $R \in \mathcal{R}$, we assume that there is a set of parameterized skills $\Pi^R=\{\pi^R_i\}_{i=1}^{N}$.
Each skill receives a parameter $x \in \mathcal{X}_i$, where $\mathcal{X}_i$ is the parameter space of skill $i$. 
The skill $\pi^R_i(x)$ generates a squence of low-level actions $(a_1, \cdots, a_t, \cdots), a_t\in\mathcal{A}$. A parameterized skill can be moving the robotic arm's end effector to a targeted position $x$.
The high-level controller outputs $(H, X)$, where $H = (h_1, \dots, h_k, \dots)$ is the skill sequence, $h_k \in [N]$ is the skill at plan step $k$, $X = (x_{h_1}^{(1)}, \dots, x_{h_k}^{(k)}, \dots)$ is the corresponding parameter sequence, and $x_{h_k}^{(k)} \in \mathcal{X}_{h_k}$ denotes the parameter for skill $h_k$.
Given a language description $L$ for the tool-use tasks, our goal is to generate a code $\tau ((H, X), \Pi, L)$ that can solve the task by calling a sequence of parameterized skills meanwhile providing their parameters.
Considering the feasible solution $\tau$ may not be unique and potentially involve manipulating different numbers of objects as tools, besides task completion, we also desire a simple plan that interacts with a minimal number of objects in the scene.

\vspace{-0.05in}
\subsection{\method: Creative Robot Tool Use with Large Language Models}
\label{sec:method_details}
\vspace{-0.1in}

We propose \method, that solves hybrid discrete-continuous planning problems with constraints through creative tool use in a zero-shot manner.
\method takes \textit{natural language instructions} as inputs making it user-friendly, and outputs \textit{executable codes} calling the robot's parameterized skills. 
\method maintains a hierarchical structure consisting of four key components (Fig.~\ref{fig:method}), including an \textit{Analyzer}, a \textit{Planner}, a \textit{Calculator} and a \textit{Coder}, each is a LLM handling one functionality.

\begin{wrapfigure}{r}{.3\textwidth}
  \vspace{-0.5cm}
  \begin{minipage}{\linewidth}
  \centering
  \includegraphics[width=1\linewidth]{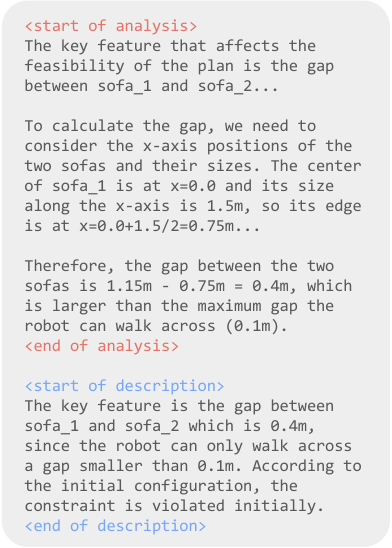}
  \vspace{-0.6cm}
  \caption{\small Analyzer output.}
  \label{fig:analyzer_outout}
  \end{minipage}
\vspace{-0.5cm}
\end{wrapfigure}

\vspace{-0.05in}
\subsubsection{Analyzer}
\vspace{-0.05in}

Humans can clearly identify crucial concepts that will affect the task plan \citep{weng2023towards}. For instance, when placing a book on a bookshelf, we use the book's dimensions, available shelf space, and slot height to determine whether the task is feasible. \textit{Can we endow robots with such reasoning capability to identify key concepts before detailed planning?}
We seek to answer the question by utilizing LLMs, which store a wealth of knowledge about objects' physical and geometric properties and human common sense.
We propose the \textit{Analyzer}, powered by LLMs, which extract the key concepts and their values that are crucial to determine task feasibility. 
\textit{Analyzer} is fed with a prompt that structures its response in two segments: an analysis section elucidating its thinking process and a description section listing the key concepts alongside their values and the related constraint. An example output of \textit{Analyzer} is shown in Fig.~\ref{fig:analyzer_outout}. We add the content in the description section of the \textit{Analyzer} output to the original descriptions $L$ to construct the key concept augmented description $L^*$ for downstream modules. 

It's worth noting that the LLMs' internalized prior knowledge autonomously determines the selection of these key concepts, and there is no prerequisite to delineating a predefined set of concepts.
Moreover, \textit{Analyzer} can extract explicit concepts provided in the description in $L$, such as the objects' positions and related workspace ranges, and implicit concepts that require calculations based on provided numerical information, such as the gap width between two objects as in Fig.~\ref{fig:analyzer_outout}.

\begin{wrapfigure}{r}{.5\textwidth}
  \vspace{-0.5cm}
  \begin{minipage}{\linewidth}
  \centering
  \includegraphics[width=1\linewidth]{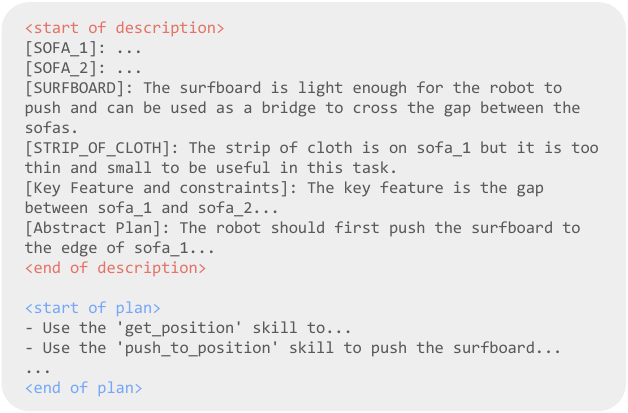}
  \vspace{-0.6cm}
  \caption{\small Planner output.}
  \label{fig:planner_outout}
  \end{minipage}
\vspace{-0.4cm}
\end{wrapfigure}

\vspace{-0.05in}

\subsubsection{Planner}
\vspace{-0.05in}

Motivated by the strong task decomposition capability of LLMs \citep{ahn2022can, huang2022language}, we propose to use an LLM as a \textit{Planner} to generate a plan skeleton $H$ based on the key concept augmented language description $L^*$.
The response of the \textit{Planner} contains a description section and a plan section, as shown in Fig.~\ref{fig:planner_outout}. 
\textit{Planner} first describes each object's properties and possible roles in finishing the task, showing the reasoning process and constructing an abstract plan in the description section. 
\textit{Planner} then generates a detailed plan skeleton based on the parameterized skills in the plan section.
We provide \textit{Planner} with a prompt describing each parameterized skill, formats of the response, and rules for the two sections.

We observe that \textit{Planner} can automatically generate a plan by utilizing objects within the environment as intermediate tools to complete a task, with the help of the key concept augmentation. 
Examples include ``using a box as a stepping stone" or ``using a hammer to drag the milk carton in the workspace."
\textit{Planner} can discover functionalities of the objects beyond their standard affordances, demonstrating the \textit{creative tool-use capability} by reasoning over the objects' physical and geometric properties.
In addition, \textit{Planner} can generate a long-horizon plan, especially in handling tasks requiring multiple tools sequentially. For instance, it can generate a plan consisting of 15 plan steps for the ``Can-Grasping" task as in Fig.~\ref{fig:tool_use_benchmark}.

\vspace{-0.05in}
\subsubsection{Calculator}
\vspace{-0.05in}
Existing literature shows that LLMs' performance tends to decline when they operate across varied levels of abstractions~\citep{liang2023code, yu2023language}. 
Inspired by these findings, we introduce a \textit{Calculator}, a standalone LLM for calculating the desired parameters for the parameterized low-level skill at each plan step, denoted as $X$ in Sec.~\ref{sec:problem_formulation}.
\textit{Calculator} processes both the key concept augmented description $L^*$ and the \textit{Planner}-generated plan skeleton $H$ for its calculation. 
Similar to the \textit{Analyzer} and \textit{Planner}, it generates a response with two sections, including a description section showing its calculation process and an answer section containing the numerical values. 
These numerical outcomes, representing target positions, are then integrated into each corresponding step of the plan skeleton. 
We provide \textit{Calculator} with multiple exemplars and rules to help deduce target positions.
\textit{Calculator} can generate navigation target positions for quadrupedal robots and push offsets for robotic arms to manipulate objects.

\vspace{-0.05in}
\subsubsection{Coder}
\vspace{-0.05in}
Finally, we introduce a \textit{Coder} module that transforms the plan $(H,X)$ into an executable code script $\tau$ that invokes robot low-level skills, perception APIs, and built-in Python libraries to interact with the environment. 
We provide \textit{Coder} with the definitions of each low-level skill and template-related rules. 
Although \textit{Coder} generates the script in an open-loop manner, the produced code is inherently designed to receive feedback from the environment. It achieves this through built-in skills like ``\texttt{get\_position}", granting a certain level of responsiveness to environmental changes.

\vspace{-0.05in}
\section{Creative Robot Tool Use Benchmark}
\label{sec:test_benchmark}
\vspace{-0.1in}

\begin{figure}[t]
    \vspace{-0.5cm}
    \centering
    \begin{subfigure}[b]{0.325\textwidth}
        \includegraphics[width=\linewidth]{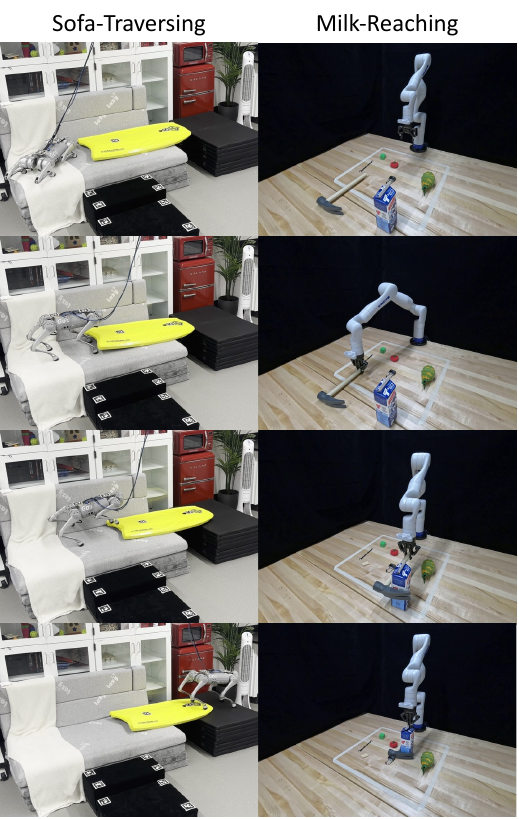}
        \caption{Tool Selection}
        \label{fig:tool_select}
    \end{subfigure}
    \hfill
    \begin{subfigure}[b]{0.325\textwidth}
        \includegraphics[width=\linewidth]{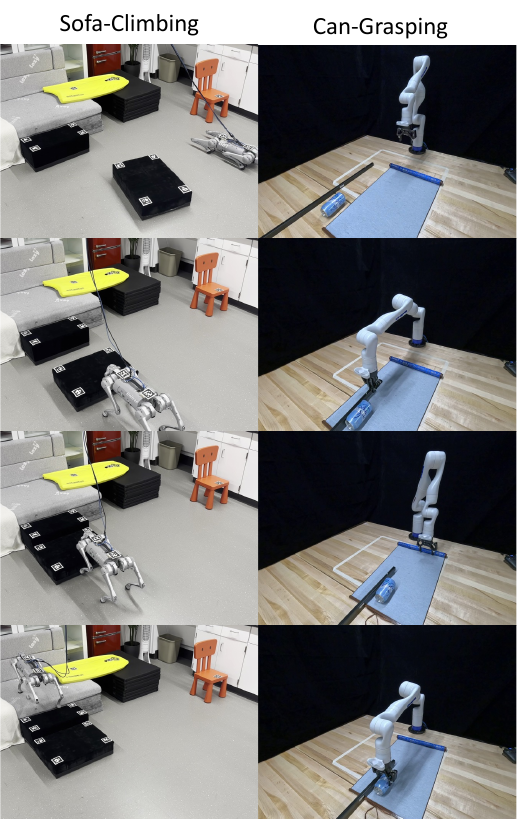}
        \caption{Sequential Tool Use}
        \label{fig:seq_tool_use}
    \end{subfigure}
    \hfill
    \begin{subfigure}[b]{0.325\textwidth}
        \includegraphics[width=\linewidth]{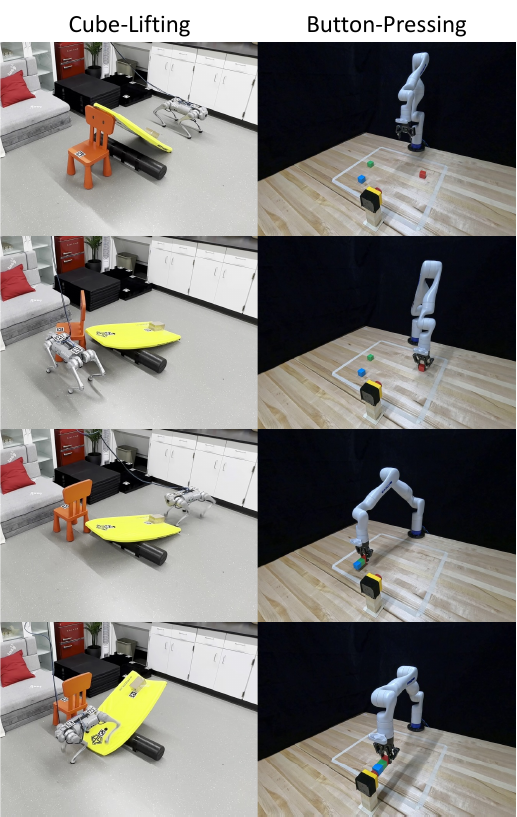}
        \caption{Tool Manufacturing}
        \label{fig:tool_manufact}
    \end{subfigure}
\vspace{-0.1in}
\caption{\small Visualization of \method's creative tool-use behaviors. (a) Tool selection. The quadrupedal robot needs to select the surfboard over a strip of cloth and push it to bridge the gap (\textbf{Sofa-Traversing}). The robotic arm needs to choose the hammer among many options and use it as a hook to pull the milk carton into the workspace (\textbf{Milk-Reaching}). (b) Sequential tool use. The quadrupedal robot needs to push a small box against a large box adjacent to the sofa and use the small box as the initial stepstone and the large box as the intermediate stepstone to climb onto the sofa (\textbf{Sofa-Climbing}). The robotic arm needs to pick up a stick, push a can onto a strip of paper, and then pull the paper closer with the can on it (\textbf{Can-Grasping}). (c) Tool manufacturing. The quadrupedal needs to identify the hidden lever structure in the environment and push away a chair supporting one end of the lever so that it can activate the lever arm and lift a heavy cube (\textbf{Cube-Lifting}). The robotic arm needs to assemble magnetic blocks to create a stick to press a button outside its workspace (\textbf{Button-Pressing}).}
\vspace{-0.2in}
\label{fig:tool_use_benchmark}
\end{figure}

Unlike conventional tool use, creative tool use, also termed as flexible tool use, is recognized by some as an indication of advanced intelligence, denoting animals' explicit reasoning about tool applications contingent on context~\citep{call_2013}. From a general problem-solving perspective, \cite{fitzgerald2021modeling} further characterized human creativity attributes, including improvisation in the absence of typical tools, use of tools in novel ways, and design of innovative tools tailored for new tasks.
In this work, we aim to explore three challenging categories of \textit{creative tool use} for robots: \textbf{tool selection}, \textbf{sequential tool use}, and \textbf{tool manufacturing}~\citep{qin2023robot}. We design six tasks for two different robot embodiments: a quadrupedal robot and a robotic arm. The details of each task are as shown in Fig.~\ref{fig:tool_use_benchmark} and Sec.~\ref{sec:appendix_task_description}, and the violated constraints of each task are listed in Tab.~\ref{tab:acc_key_feature}. 
\begin{itemize}[leftmargin=0.5cm]
    \itemsep-0.1em
    \item \textbf{Tool selection} (Sofa-Traversing and Milk-Reaching) requires the reasoning capability to choose the most appropriate tools among multiple options. It demands a broad understanding of object attributes such as size, material, and shape, as well as the ability to analyze the relationship between these properties and the intended objective.
    \item \textbf{Sequential tool use} (Sofa-Climbing and Can-Grasping) entails utilizing a series of tools in a specific order to reach a desired goal. Its complexity arises from the need for long-horizon planning to determine the best sequence for tool use, with successful completion depending on the accuracy of each step in the plan.
    \item \textbf{Tool manufacturing} (Cube-Lifting and Button-Pressing) involves accomplishing tasks by crafting tools from available materials or adapting existing ones. This procedure requires the robot to discern implicit connections among objects and assemble components through manipulation.
\end{itemize}

\vspace{-0.05in}
\section{Experiment Results}
\label{sec:exp_results}
\vspace{-0.1in}

We aim to investigate whether \method possesses various types of creative tool-use capabilities by evaluating it on the benchmark outlined in Sec.~\ref{sec:test_benchmark}. 
We build both simulation and real-world platforms detailed in Sec.~\ref{sec:exp_setup} and compare \method with four baselines described in Sec.~\ref{sec:baselines}.
We measure the task success rates to understand the performance of \method in Sec.~\ref{sec:tool_use_capability} and analyze the effect of \method's modules through error breakdown in Sec.~\ref{sec:error_breakdown}. 
We then dive deeper into the role of \textit{Analyzer} and show that it enables discriminative creative tool-use behaviors in Sec~\ref{sec:discriminative_tool_use}.

\subsection{Experiment Setup}
\vspace{-0.05in}
\label{sec:exp_setup}
\textbf{Robotic Arm.} We test \method with a Kinova Gen3 robotic arm (details in Sec.~\ref{sec:appendix_arm}).
In simulation, we build tasks based on robosuite \citep{zhu2020robosuite} and assume known object positions and sizes. 
In real-world experiments, we employ OWL-ViT \citep{minderer2022simple} to obtain 2D locations and bounding boxes for each object. 
In both platforms, the robot maintains a skill set as [``\texttt{get\_position}", ``\texttt{get\_size}", ``\texttt{open\_gripper}", ``\texttt{close\_gripper}", ``\texttt{move\_to\_position}"]. 
Note that we use skills without explicitly listing the object-centric movements caused by the ``\texttt{move\_to\_position}" skill, such as pushing or picking.

\textbf{Quadrupedal Robot.} We test \method with a Unitree Go1 quadrupedal robot (details in Sec.~\ref{sec:appendix_quadruped_real}).
The simulation experiments for quadrupedal robots are evaluated based on the generated code and through human evaluations. 
In real-world experiments, considering the relatively large workspace compared with the tabletop setting when experimenting with the robotic arm, we use AprilTags~\citep{olson2011apriltag} affixed to each object in real-world experiments to get the object's positions.
Each skill in real-world experiments is equipped with skill-specific motion planners to generate smooth and collision-free velocity commands for different walking modes of Go1. 
For both simulation and real-world experiments, the quadrupedal robot's skill set is [``\texttt{get\_position}", ``\texttt{get\_size}", ``\texttt{walk\_to\_position}", ``\texttt{climb\_to\_position}", ``\texttt{push\_to\_position}"].

\vspace{-0.05in}
\subsection{Baselines}
\label{sec:baselines}
\vspace{-0.05in}
We compare \method with four baselines, including one variant of Code-as-Policies~\citep{liang2023code} and three variants of our proposed \method.
\vspace{-0.2cm}
\begin{itemize}[leftmargin=0.5cm]
    \itemsep-0.1em
    \item \textbf{Coder.} It takes the natural language instruction as input and directly outputs executable code. It is a variant motivated by Code-as-Policies~\citep{liang2023code}. This baseline demonstrates the combinatorial effect of the other three modules in \method.
    \item \textbf{Planner-Coder.} It removes the \textit{Analyzer} and the \textit{Calculator} in \method. This baseline demonstrates the combinatorial effect of the \textit{Analyzer} and the \textit{Calculator} modules.
    \item \textbf{\method without Analyzer.} The \textit{Planner} directly takes the language instruction as input. This baseline reveals the effect of the \textit{Analyzer} in the downstream planning.
    \item \textbf{\method without Calculator.} The \textit{Coder} directly takes the response of the \textit{Planner}. This baseline demonstrates the effect of the \textit{Calculator} module.
\end{itemize}
\vspace{-0.2cm}
We evaluate \method both in simulation and in the real world while only evaluating of baselines in simulation given their relatively low success rates in simulation. \method's prompts are in Sec.~\ref{sec:appendix_prompts}.

\begin{table}[t]
    \centering
    \vspace{-0.5cm}
    \caption{\small Success rates of \method and baselines. Each value is averaged across 10 runs. All methods except for \textbf{\method (Real World)} are evaluated in simulation.}
    \vspace{-0.1in}
    \resizebox{0.9\textwidth}{!}{
    \begin{tabular}{lccccccc} 
    \toprule
        & \textbf{Milk-} & \textbf{Can-} & \textbf{Button-} & \textbf{Sofa-} & \textbf{Sofa-} & \textbf{Cube-} & \multirow{2}{*}{\textbf{Average}}\\ 
        & \textbf{Reaching} & \textbf{Grasping} & \textbf{Pressing} & \textbf{Traversing} & \textbf{Climbing} & \textbf{Lifting} & \\ \midrule
        \textbf{\method}                & \textbf{0.9} & \textbf{0.7} & \textbf{0.8} & \textbf{1.0} & \textbf{1.0} & \textbf{0.8} & \textbf{0.87}\\
        \textbf{\method w/o Analyzer}           & 0.0 & 0.4 & 0.2 & \textbf{1.0} & 0.7 & 0.2 & 0.42\\
        \textbf{\method w/o Calculator}            & 0.0 & 0.1 & \textbf{0.8} & 0.3 & 0.0 & 0.3 & 0.25\\
        \textbf{Planner-Coder} & 0.0 & 0.2 & 0.5 & 0.1 & 0.0 & 0.4 & 0.20\\
        \textbf{Coder}                & 0.0 & 0.0 & 0.0 & 0.0 & 0.0 & 0.4 & 0.07\\ \midrule
        \textbf{\method (Real World)} & 0.7 & 0.7 & 0.8 & 0.7 & 0.8 & 0.9 & 0.77 \\
    \bottomrule
    \end{tabular}
    }
    \label{tab:success_rate}
\end{table}

\begin{table}[t]
\vspace{-0.3cm}
    \centering
    \caption{\small \method's proposed key concept accuracy in simulation. Each value is averaged across 10 runs.}
    \vspace{-0.1in}
    \resizebox{0.7\textwidth}{!}{
    \begin{tabular}{llc} 
    \toprule
    & \textbf{Key Concept and Violated Constraints} & \textbf{Accuracy} \\ \midrule
    \textbf{Milk-Reaching} & Milk's position is out of robot workspace. & 1.0 \\
    \textbf{Can-Grasping} & Can's position is out of robot workspace. & 1.0 \\
    \textbf{Button-Pressing} & Button's position is out of robot workspace. & 0.9 \\
    \textbf{Sofa-Traversing} & Gap's width is out of robot's walking capability. & 1.0 \\
    \textbf{Sofa-Climbing} & Sofa's height is out of robot's climbing capability.& 0.8\\
    \textbf{Cube-Lifting} & Cube's weight is out of robot's pushing capability. & 1.0\\
    \bottomrule
    \end{tabular}
    }
    \label{tab:acc_key_feature}
\end{table}

\vspace{-0.05in}
\subsection{Can \method achieve creative tool use?}
\label{sec:tool_use_capability}
\vspace{-0.05in}
We present the quantitative success rates of \method and baselines in Tab.~\ref{tab:success_rate} and real-world qualitative visualizations of \method in Fig.~\ref{fig:tool_use_benchmark}. 
\method consistently achieves success rates that are either comparable to or exceed those of the baselines across six tasks in simulation.
\method's performance in the real world drops by 0.1 in comparison to the simulation result, mainly due to the perception errors and execution errors associated with parameterized skills, such as the quadrupedal robot falling down the soft sofa. Nonetheless, RoboTool (Real World) still surpasses the simulated performance of all baselines.
Considering that the tasks in Sec.~\ref{sec:test_benchmark} are infeasible without manipulating objects as tools, we show that \method can successfully enable robot tool-use behaviors.
Moreover, as visualized in Fig.~\ref{fig:tool_use_benchmark}, \method transcends the standard functionalities of objects and creatively capitalizes on their physical and geometric properties, including materials, shapes, and sizes. Here are some highlights of the creative tool-use behaviors.

\textbf{Piror Knowledge.} In the Milk-Reaching task (Fig.~\ref{fig:tool_select}), \method leverages LLM's prior knowledge about all the available objects' shapes and thus improvisationally uses the hammer as an L-shape handle to pull the milk carton into the workspace. 

\textbf{Long-horizon Planning.} In the Can-Grasping task (Fig.~\ref{fig:seq_tool_use}), \method sequentially uses the stick to push the can onto the scroll and then drag the scroll into the workspace with the can on it. This reveals \method's long-horizon planning capability by generating a plan with as many as 15 steps. 

\textbf{Hidden Mechanism Identification.} In the Cube-Lifting task with the quadrupedal robot (Fig.~\ref{fig:tool_manufact}), \method identifies the potential existence of a mechanism consisting of the yoga roller as the fulcrum and the surfboard as the lever. \method first constructs the lever by pushing the chair away, then activates the lever by walking to one end of the lever, and finally lifts the cube. It illustrates that \method can not only fabricate a tool from available objects but also utilize the newly created tool.

\method without Analyzer performs worse than \method while better than \method without Calculator. 
Moreover, they perform better than baselines lacking \textit{Analyzer} and \textit{Calculator}, including Planner-Coder and Coder. 
These observations show that both \textit{Analyzer} and \textit{Calculator} are critical in achieving high success rates, and \textit{Calculator} plays a more important role in tasks that require accurate positional offsets such as Milk-Reaching, Can-Grasping and Sofa-Climbing.

\vspace{-0.05in}
\subsection{Error Breakdown}
\label{sec:error_breakdown}
\begin{figure}[t]
\vspace{-0.9cm}
\begin{subfigure}[b]{0.49\textwidth}
  \raisebox{0.09cm}{\includegraphics[width=1.0\linewidth]{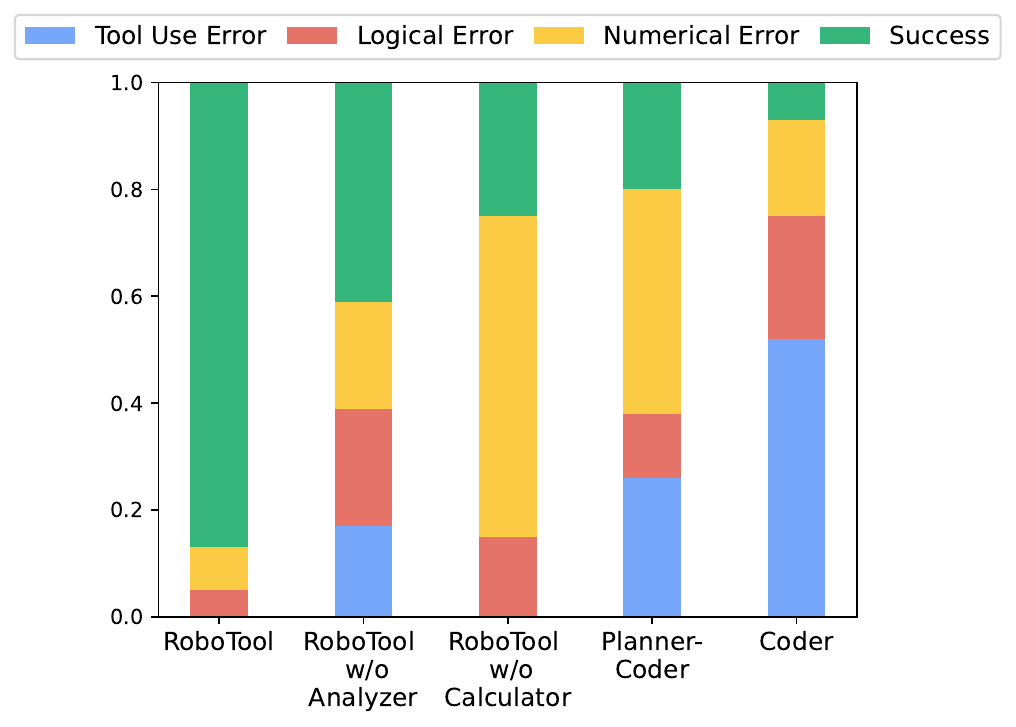}}
  \vspace{-0.65cm}
  \caption{Error breakdown}
  \label{fig:error_bar}
\end{subfigure}
\hfill
\begin{subfigure}[b]{0.49\textwidth}
    \raisebox{0.0cm}{\includegraphics[width=1.0\linewidth]{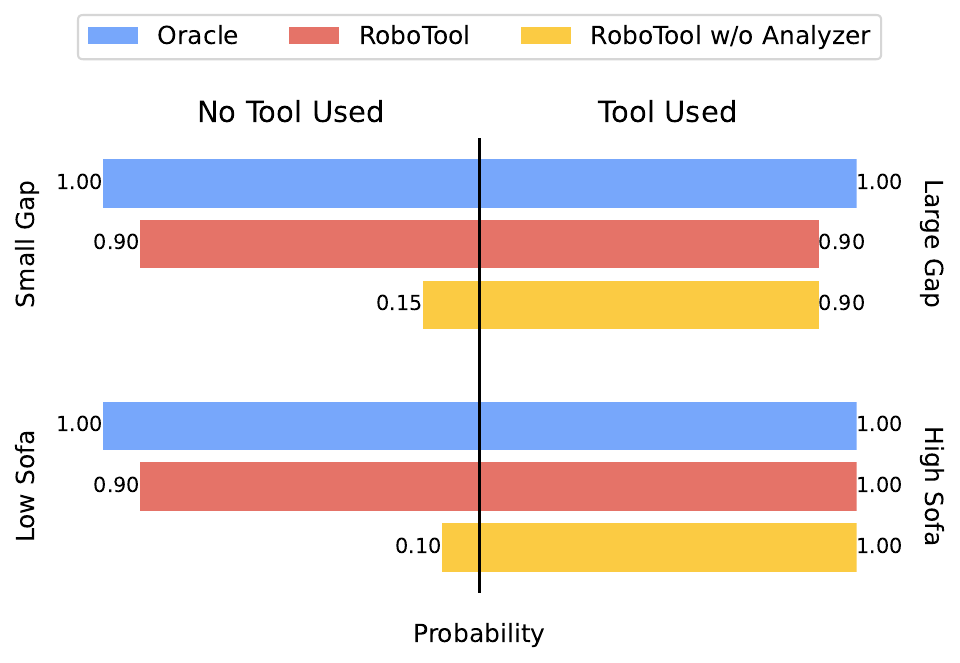}}
    \caption{Discriminative tool-use behavior}
\end{subfigure}
\vspace{-0.1in}
\caption{\small (a) Error breakdown of \method and baselines. (b) Discriminative tool-use behavior is enabled by \textit{Analyzer}'s explicit reasoning about the key concepts.
}
\label{fig:discriminative_tool_use}
\vspace{-0.2in}
\end{figure}

\vspace{-0.05in}
We further analyze what causes the failure of \method and baselines based on simulation experiments. We define three types of errors: tool-use error, logical error, and numerical error.
The tool-use error indicates whether the correct tool is used.
The logical error mainly focuses on planning error, such as using tools in the wrong order or ignoring the constraints provided. 
The numerical error includes calculating the wrong target positions or adding incorrect offsets. 
We show the error breakdown averaged across six tasks in Fig.~\ref{fig:error_bar}. 
The results show that the \textit{Analyzer} helps reduce the tool-use error when comparing \method and \method without Analyzer. 
\textit{Calculator} significantly reduces the numerical error when comparing \method, \method without Calculator and Planner-Coder. We provide per-task error breakdown results in Sec.~\ref{sec:appendix_error_bar}.

\vspace{-0.05in}
\subsection{How does Analyzer Affect the Tool-Use Capability?}
\label{sec:discriminative_tool_use}
\vspace{-0.07in}
\textbf{Key Concept Identification Accuracy.} We show the accuracy of the proposed key concept in Tab.~\ref{tab:acc_key_feature}, based on whether the \textit{Analyser} correctly returns the key concept, the value of the key concept, and the related constraint. The target responses are provided by human. The results show that \textit{Analyzer} could correctly identify the key concept that affects the plan's feasibility and accurately calculate the key concepts' value. For instance, in the Sofa-Traversing task, the key concept is identified as the distance between the boundaries of the two sofas, which is the gap width the robot needs to cover. Moreover, the \textit{Analyzer} could link the key concept with the robot's limit: the quadrupedal robot can only walk across a gap over 0.1m.

\textbf{Discriminative Tool-use Capability.} 
Given the impressive creative tool-use capability of \method, we want to investigate further whether \method possesses the discriminative tool-use capability, which is using tools when necessary and ignoring tools when the robot can directly finish tasks without the need to manipulate other objects.
We choose Sofa-Traversing and Sofa-Climbing to test the discriminative tool-use capability.
For Sofa-Traversing, we compare the rate of tool use in scenarios with large gaps where using tools is necessary, against scenarios with small gaps that allow the robot to traverse to another sofa without using tools.
For Sofa-Climbing, we evaluate the tool-use rate in scenarios where a high-profile sofa requires the use of boxes as stepstones, in contrast to low-profile sofas, in which the robot can ascend directly without assistance.

We compare the \method with an Oracle that can derive the most efficient plan and the baseline \method without Analyzer, and present the main results in Fig.~\ref{fig:discriminative_tool_use}.
In both sets of tasks, \method tends not to use tools when unnecessary (Small Gap and Low Sofa), demonstrating more adaptive behaviors given different environment layouts.
In contrast, without the help of \textit{Analyer}, the baseline tends to use tools in all four scenarios, dominated by the prior knowledge in LLMs.
These observations show that \textit{Analyser} helps enable the discriminative tool-use behavior of \method.

\vspace{-0.12in}
\section{Conclusion}
\label{sec:conclusion}
\vspace{-0.13in}
We introduce \method, a creative robot tool user powered by LLMs that enables solving long-horizon planning problems with implicit physical constraints. 
\method contains four components: (i) an
``Analyzer” that discerns crucial task feasibility-related concepts, (ii) a ``Planner” that generates creative tool-use plans, (iii) a ``Calculator” that computes skills' parameters, and (iv) a ``Coder” that generates executable code. 
We propose a benchmark to evaluate three creative tool-use behaviors, including tool selection, sequential tool use, and tool manufacturing. 
Through evaluating on the creative tool use benchmark, we show that \method can identify the correct tool, generate precise tool-usage plans, and create novel tools to accomplish the task. We compare our method to four baseline methods and demonstrate that \method achieves superior performance when the desired tasks require precise and creative tool use.

\textbf{Limitations. }
Since we focus on the tool-use capability of LLMs at the task level in this paper, we rely on existing APIs to process visual information, such as describing the graspable points of each object and summarizing the scene. It is possible to integrate vision language models to replace the designed API to get the affordance for each object similar to VoxPoser \citep{huang2023voxposer}.  In addition, we highlight that the proposed method serves as a planner, specializing in executable plan creation, not an execution framework.
Reactive execution with a feedback loop could be achieved by integrating hybrid shooting and greedy search into our method, such as in \cite{lin2023text2motion}.

\subsubsection*{Acknowledgments}
We would like to thank Jacky Liang for the feedback and suggestions, and Yuyou Zhang, Yikai Wang, Changyi Lin for helping set up real-world experiments.

\newpage
\bibliography{main}
\bibliographystyle{iclr2024_conference}

\newpage
\appendix
\section{Additional Experiment Results}
\label{sec:appendix_error_bar}
\begin{figure}[h]

\begin{subfigure}[b]{0.49\textwidth}
  \raisebox{0.0cm}{\includegraphics[width=1.0\linewidth]{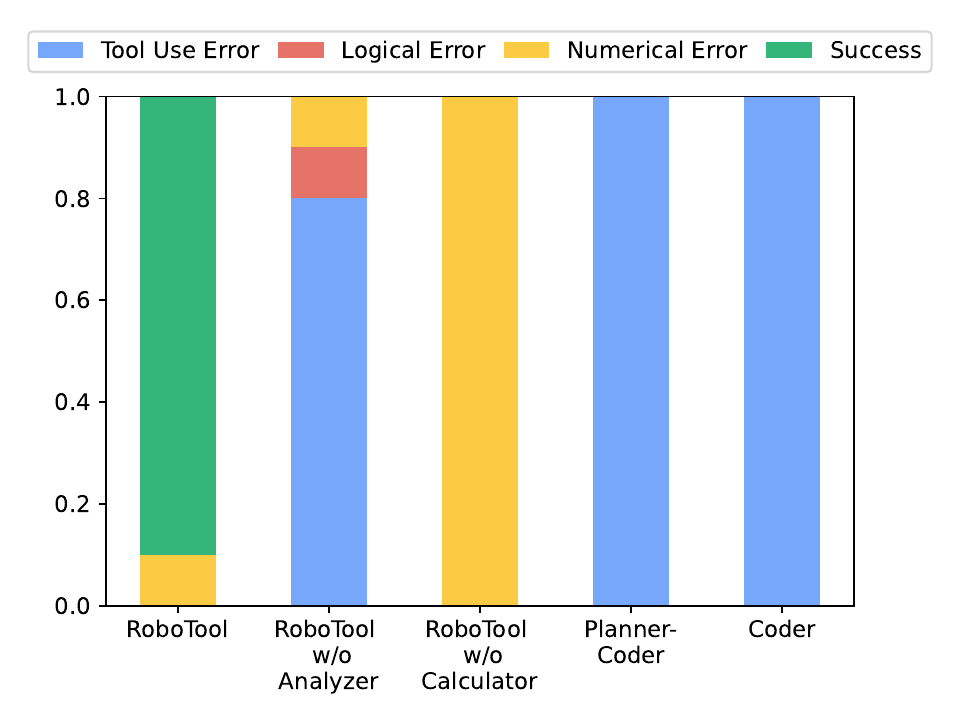}}
  \vspace{-0.5cm}
  \caption{Milk-Reaching}
\end{subfigure}
\hfill
\begin{subfigure}[b]{0.49\textwidth}
  \raisebox{0.0cm}{\includegraphics[width=1.0\linewidth]{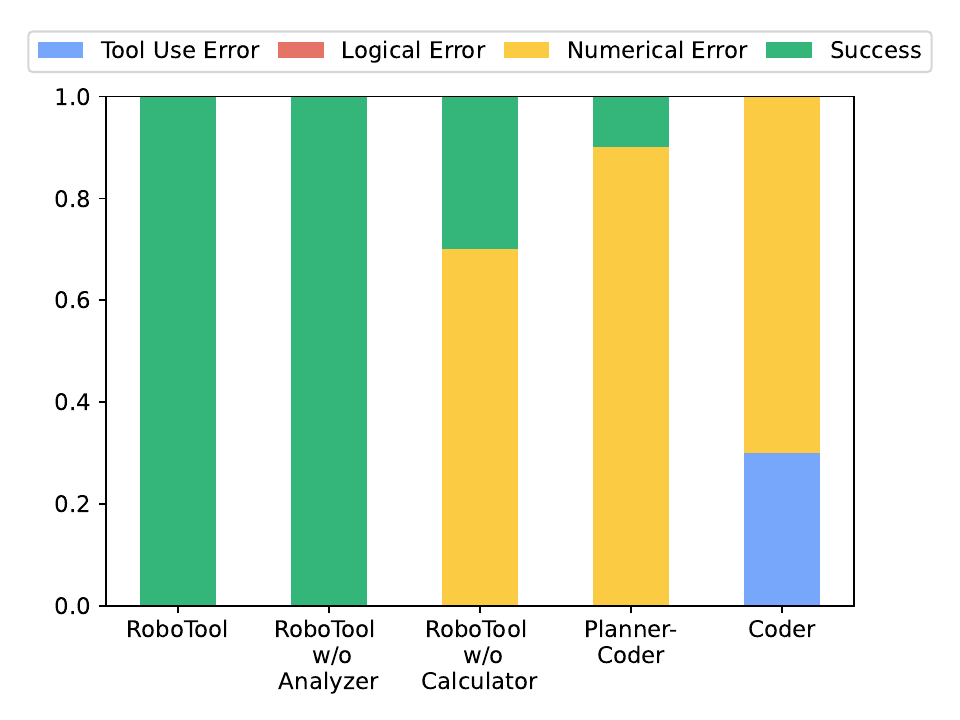}}
  \vspace{-0.5cm}
  \caption{Sofa-Traversing}
\end{subfigure}
\hfill
\begin{subfigure}[b]{0.49\textwidth}
  \raisebox{0.0cm}{\includegraphics[width=1.0\linewidth]{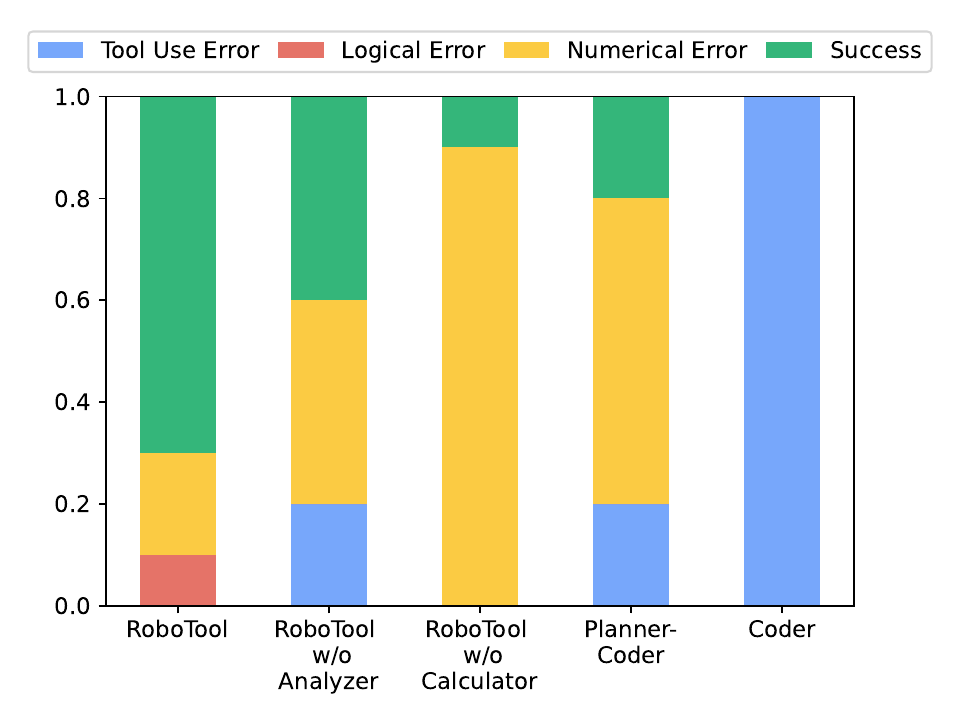}}
  \vspace{-0.5cm}
  \caption{Can-Grasping}
\end{subfigure}
\hfill
\begin{subfigure}[b]{0.49\textwidth}
  \raisebox{0.0cm}{\includegraphics[width=1.0\linewidth]{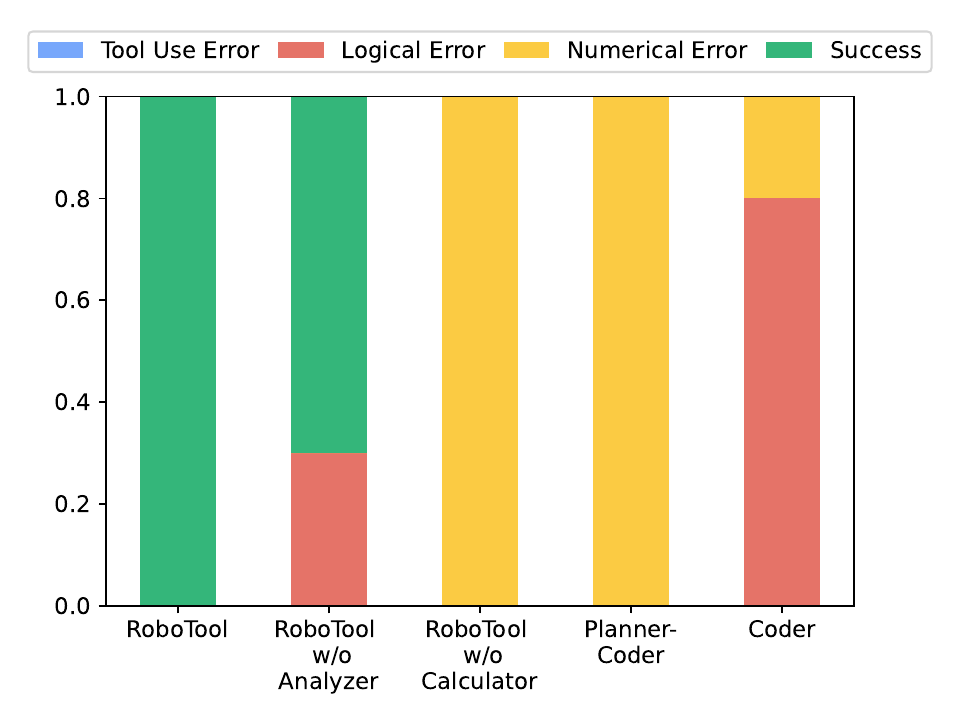}}
  \vspace{-0.5cm}
  \caption{Sofa-Climbing}
\end{subfigure}
\hfill
\begin{subfigure}[b]{0.49\textwidth}
  \raisebox{0.0cm}{\includegraphics[width=1.0\linewidth]{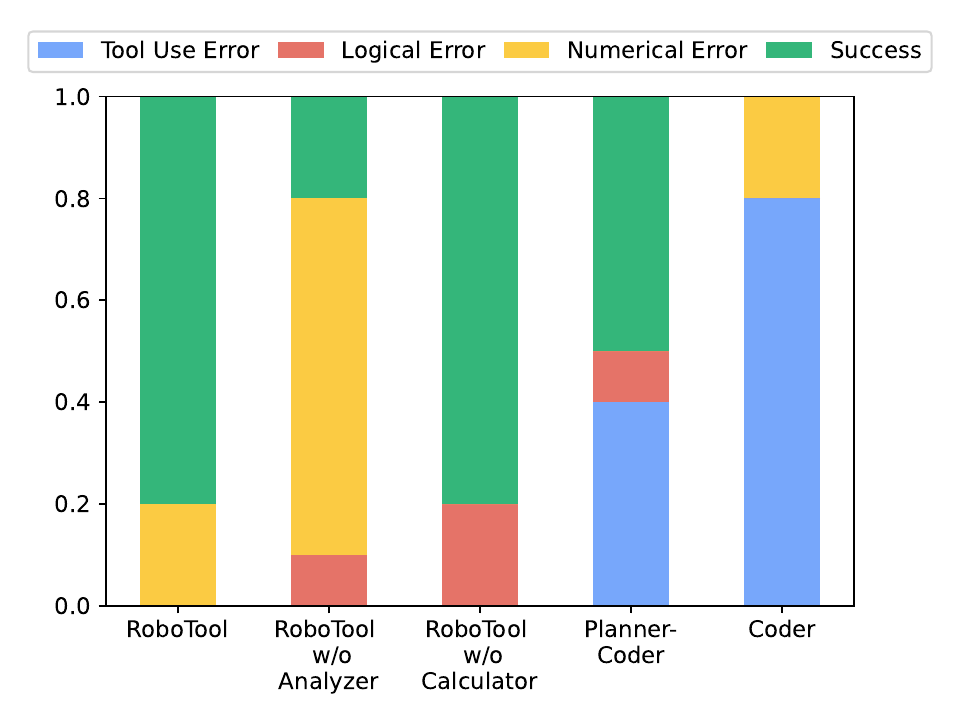}}
  \vspace{-0.5cm}
  \caption{Button-Pressing}
\end{subfigure}
\hfill
\begin{subfigure}[b]{0.49\textwidth}
  \raisebox{0.0cm}{\includegraphics[width=1.0\linewidth]{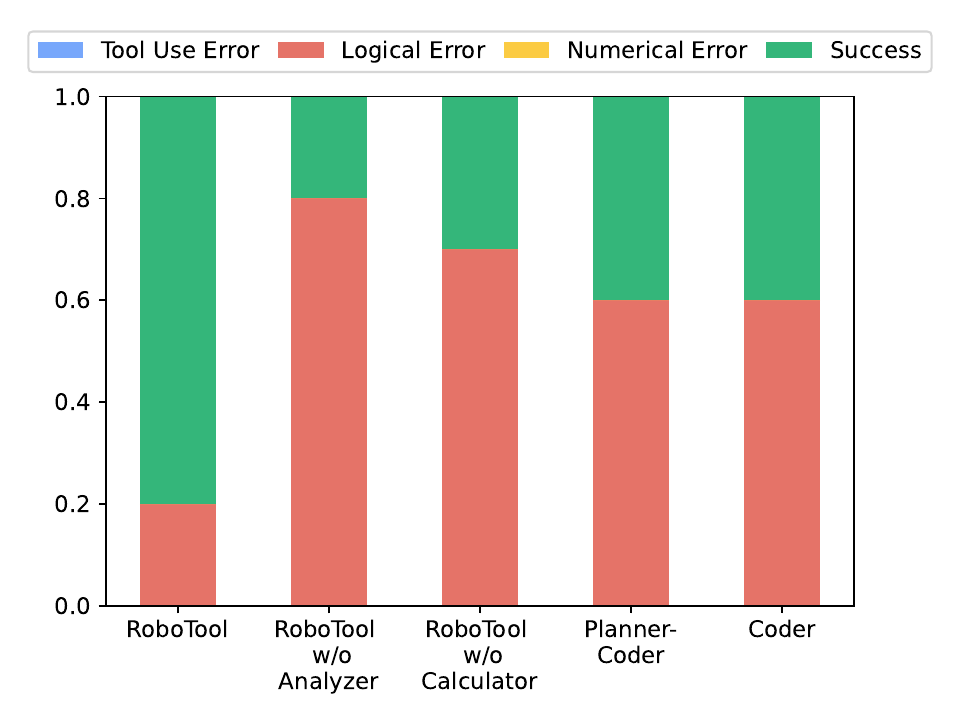}}
  \vspace{-0.5cm}
  \caption{Cube-Lifting}
\end{subfigure}
\vspace{-0.05in}
\caption{\small Error breakdown for each creative tool-use task.
}
\label{fig:appendix_error_bar}
\vspace{-0.1in}
\end{figure}

We provide additional error breakdown results in Fig.~\ref{fig:appendix_error_bar}. We observe that different modules play different roles in various tasks. 
For instance, in the Milk-Reaching task, the Planner-Coder baseline is dominated by the tool use error without knowing using the hammer as the tool to drag the milk into the workspace. In this case, \textit{Analyzer} helps reduce the tool use error significantly.
In contrast, in the Cube-Lifting task, most of the generated plans could construct a lever by pushing away the chair. However, the baselines tend to ignore the cube's weight and assume that the dropping surfboard could automatically lift the cube. In this case, the \textit{Analyzer} helps reduce the logical error. 
While in other tasks, the \textit{Calculator} becomes quite important, especially in Can-Grasping, Sofa-Climbing, and Milk-Reaching.

\section{Real-World Setup for Quadrupedal Robot} 
\label{sec:appendix_quadruped_real}

\begin{figure}[htb]
\centering
\includegraphics[width=0.9\linewidth]{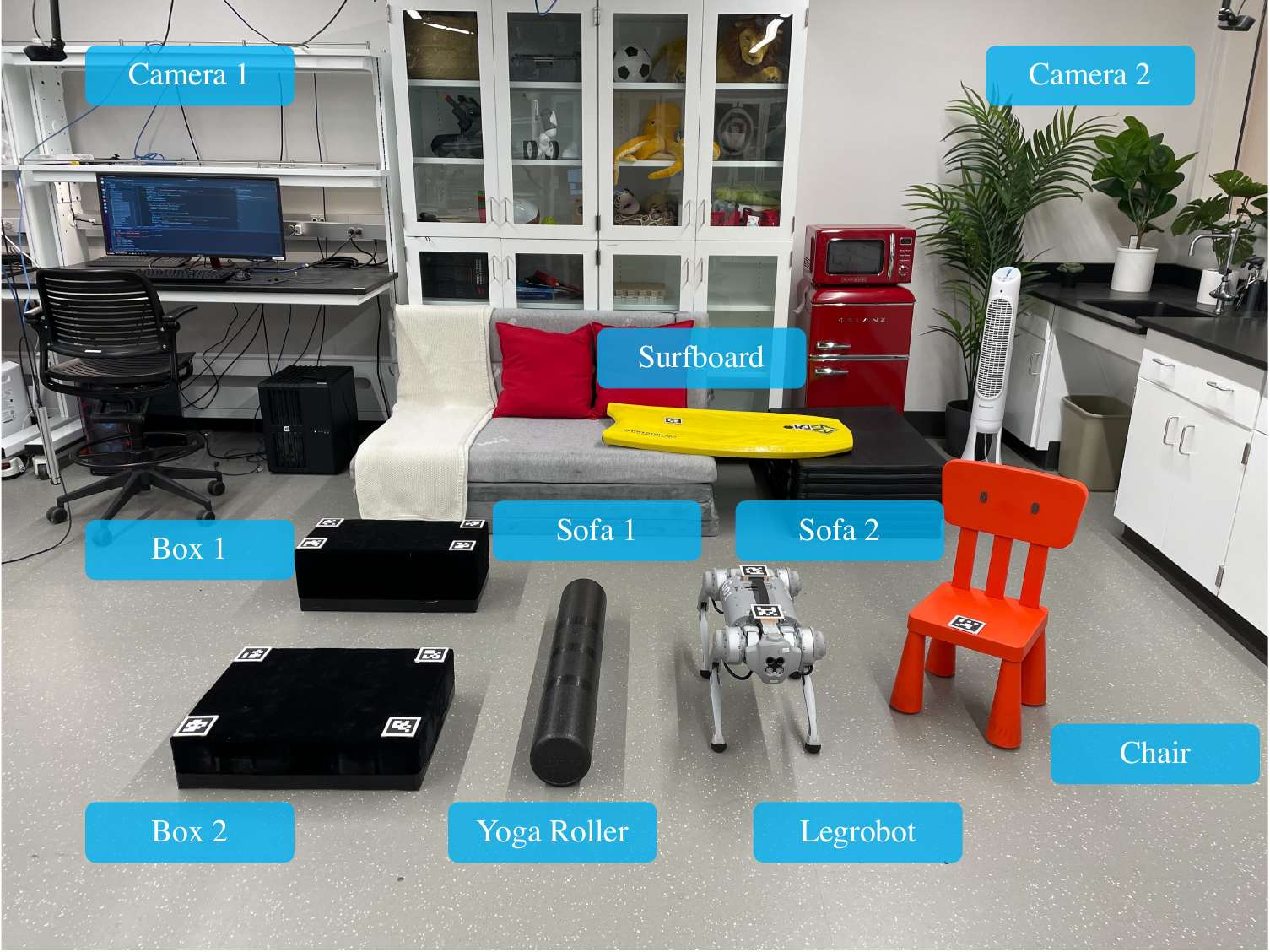}
\caption{The figures illustrate the quadrupedal robot environment setup, including object names and geometries. The image shows various objects with which the robot can interact. The above-described objects are labeled with names in the figure.}
\label{fig:legrobot_setup}
\end{figure}

In the quadrupedal robot environment setup, several objects with which the robot can interact are presented in Fig.~\ref{fig:legrobot_setup}. These include two blocks of varying heights, two sofas positioned adjacently with gaps between them, a chair, a surfboard, and a yoga roller. Additionally, two ZED2 cameras are situated at the top-left and top-right of the environment to capture the positions of the robot and other objects.
These April tags are identifiable by the two ZED 2 cameras situated at the top-left and top-right of the environment, enabling the computation of the object's position using the PnP algorithm \cite{fischler_bolles_1981}.

The quadrupedal robot possesses five skills within its skill set: \texttt{move\_to\_position}, \texttt{push\_to\_position}, \texttt{climb\_to\_position}, \texttt{get\_position}, \texttt{get\_size}.

\subsection{move\_to\_position}
\begin{figure}[htb]
\centering
\includegraphics[width=1.0\linewidth]{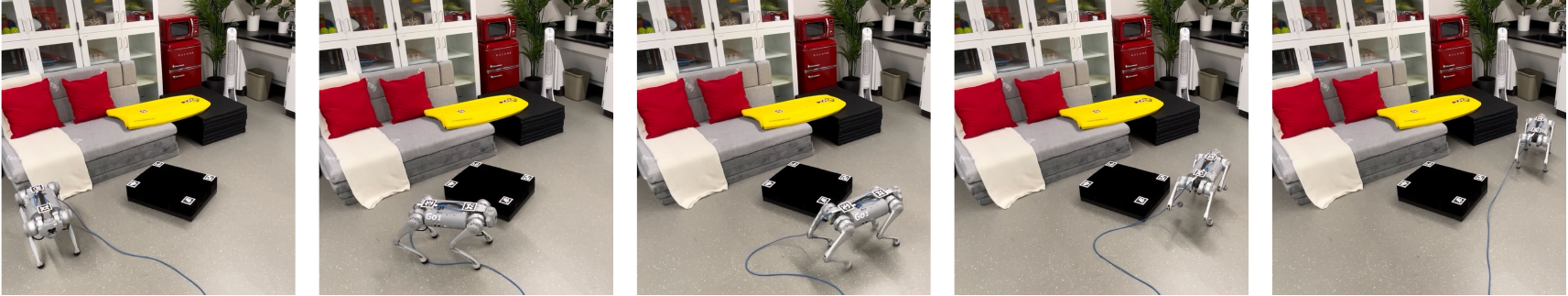}
\caption{This figure illustrates the robot's moving skill, reaching the target location while avoiding an obstacle on the path. The collision-free trajectory is generated by an informed RRT* path planner.}
\label{fig:legrobot_skill_walk}
\end{figure}

Upon invoking this skill, the quadrupedal robot navigates to the target position from its current location, avoiding obstacles present in the scene. The movement is facilitated using the built-in trot gait in continuous walking mode from Unitree. Trajectories are generated using the informed RRT* method \cite{6942976} to prevent potential collisions during trajectory planning. Fig.~\ref{fig:legrobot_skill_walk} illustrates an example of a trajectory produced by the motion planner and demonstrates the robot's movement along this path.

\subsection{push\_to\_position}
\begin{figure}[htb]
\centering
\includegraphics[width=0.8\linewidth]{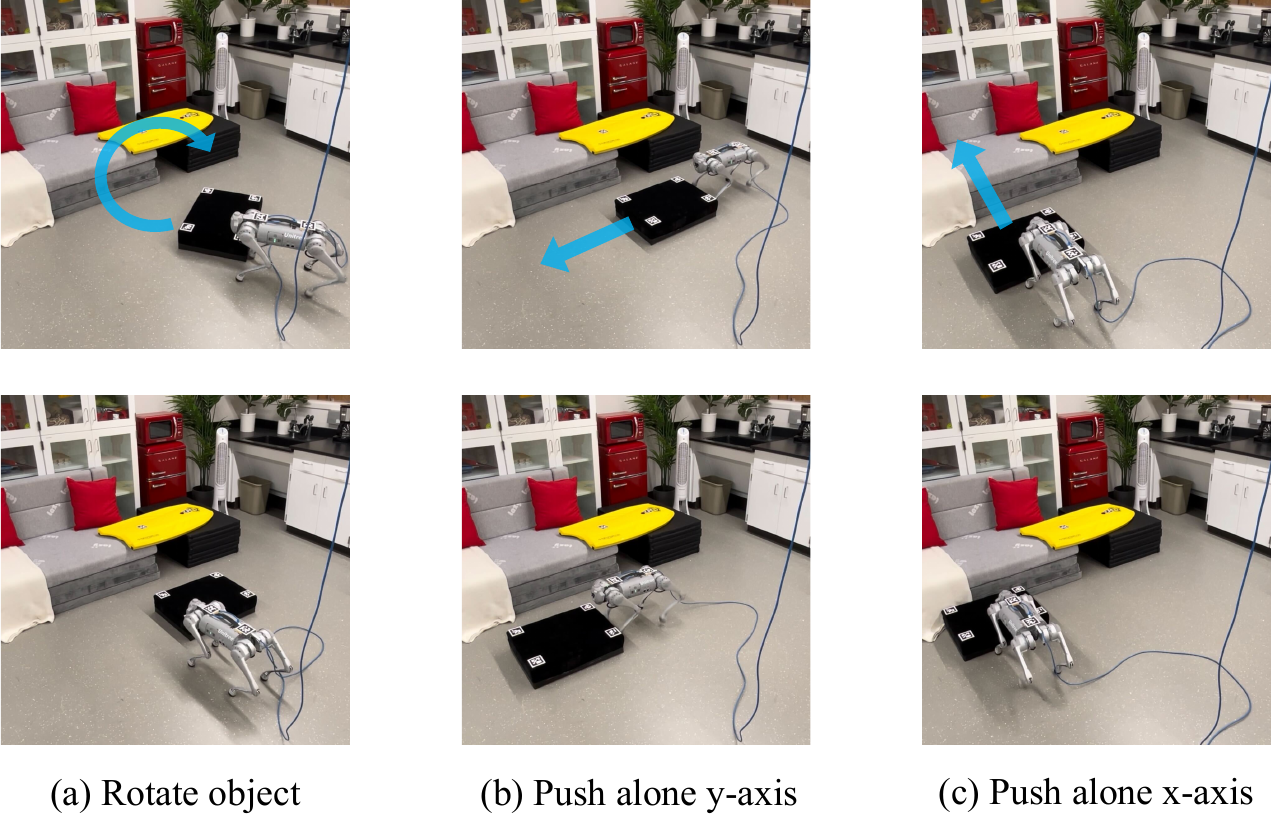}
\caption{This figure demonstrates the robot's object-pushing skill. (a) First, the robot rotates the object by pushing one corner. (b) Then, it pushes the object along the y-axis, (c) followed by the x-axis, until the object reaches its designated location.}
\label{fig:legrobot_skill_push}
\end{figure}

When this skill is called, the quadrupedal robot pushes an object to the target location following this sequence, also as demonstrated in Fig.~\ref{fig:legrobot_skill_push}:
\begin{enumerate}
    \item \textbf{Rotate Object:} The quadrupedal robot initially attempts to rotate the object until its rotation along the z-axis aligns with the target.
    \item \textbf{Push along y-axis:} The quadrupedal robot subsequently attempts to push the object along the y-axis until the object's y-position matches the target.
    \item \textbf{Push along x-axis:} Finally, the quadrupedal robot pushes the object along the x-axis until the object's x-position meets the target.
\end{enumerate}

\subsection{climb\_to\_position}
This skill enables the robot to climb to the desired location utilizing the built-in stair-climbing mode from Unitree. Path planning is disabled when this skill is invoked because the robot is able to move above obstacles. 

\begin{figure}[htb]
\centering
\includegraphics[width=1.0\linewidth]{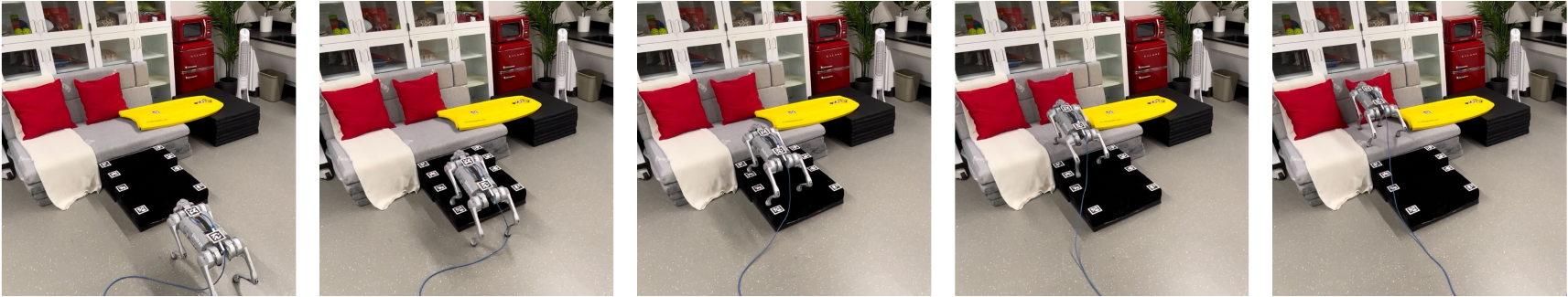}
\caption{An illustration showcasing the robot's climbing skill, the robot successfully ascends onto a sofa by navigating through box 1 and box 2 of differing heights.}
\label{fig:legrobot_skill_climb}
\end{figure}

Fig.~\ref{fig:legrobot_skill_climb} illustrates an instance where the robot climbed onto a sofa by climbing on two boxes of varying heights.

\subsection{get\_position}
The position of each object is estimated using AprilTags affixed to them. These AprilTags are identifiable by the two ZED 2 cameras, enabling the computation of the object's position using the PnP algorithm \citep{fischler_bolles_1981}.
Fig.~\ref{fig:legrobot_skill_get_position} shows the estimated positions of some objects from one camera.

\subsection{get\_size}
The bounding boxes of the objects are pre-measured and stored in a database. Each time this function is invoked, it returns the object's size based on its position and orientation.

\begin{figure}[htb]
\centering
\includegraphics[width=1.0\linewidth]{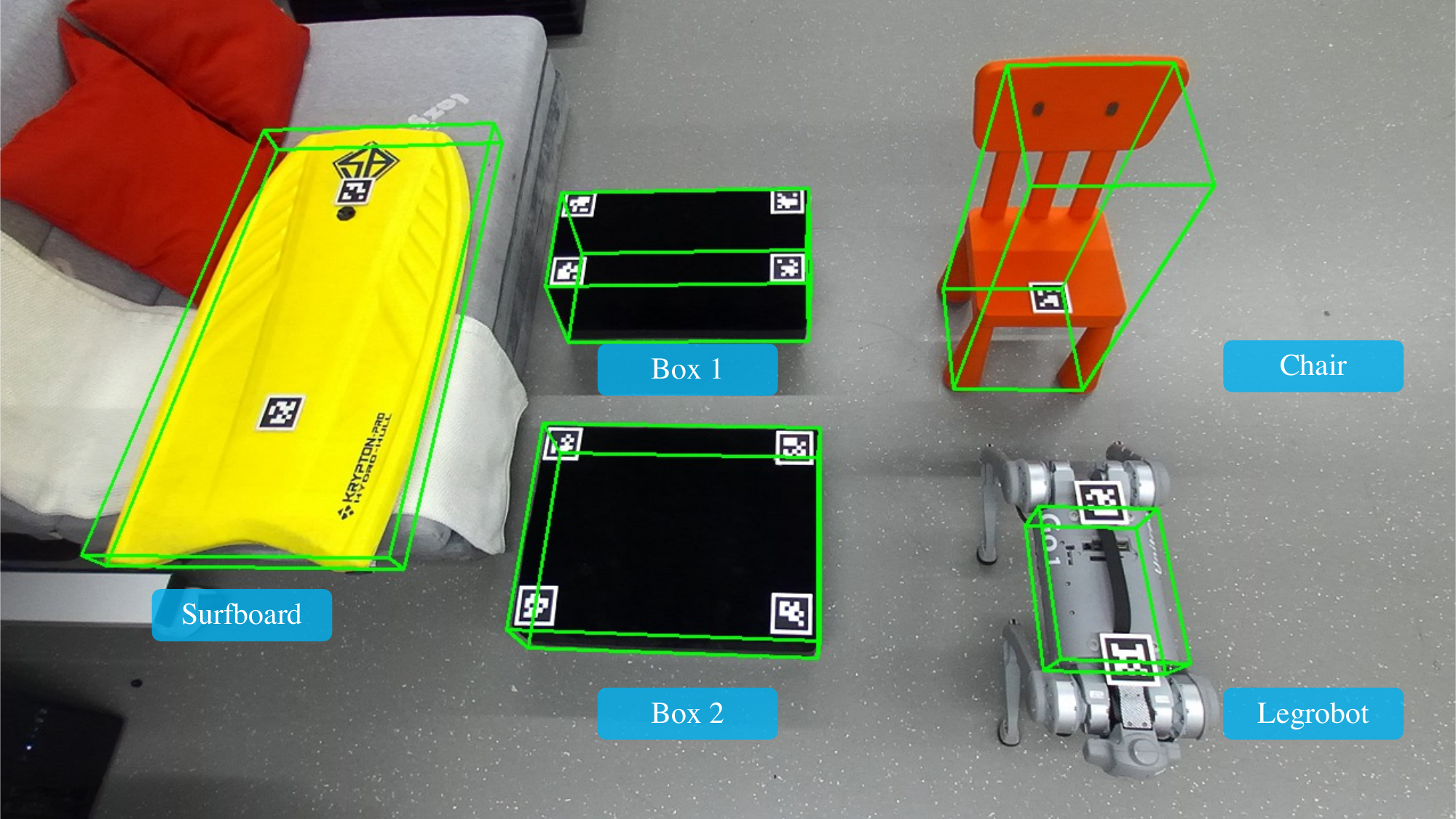}
\caption{This image illustrates the estimated positions and bounding boxes of various objects, as computed by the PnP algorithm by capturing AprilTags placed on each object. The objects shown include box 1, box 2, a surfboard, a quadrupedal robot, and a chair.}
\label{fig:legrobot_skill_get_position}
\end{figure}

Fig.~\ref{fig:legrobot_skill_get_position} illustrated some object bounding boxes estimated from one camera.

\section{Real-World Setup for the Robotic Arm} 
\label{sec:appendix_arm}

We test \method using a Kinova Gen3 Robot arm with 7 degrees of freedom and a two-fingered gripper.
In real-world experiments, we applied the OWL-ViT \cite{minderer2022simple} to obtain 2D locations and bounding boxes for each object. 
We did this by capturing a slightly tilted top-down view of the scene.
Next, we converted the coordinates of the bounded image from 2D to both world coordinates and robot coordinates.
Finally, we combined the depth information of each detected object with the transformed 2D bounding box in robot coordinates to calculate the complete 3D position and size of the objects in the scene.

We assume the graspable point of each object is given to \method. Graspable point of objects is a long-standing and active research field in robotics \citep{fang2020learning, lin2015robot, myers2015affordance, song2010learning, ek2010exploring, song2015task}. In this work, we focus on the high-level planning capability of LLMs rather than the low-level grasping policy.

In the robot arm environment setup, the tasks focus on table-top manipulations. such as Button-Pressing, Milk-Reaching, and Can-Grasping. Tasks are executed by the combination of the following skills: \texttt{move\_to\_position}, \texttt{open\_gripper}, \texttt{close\_gripper}, \texttt{get\_position}, \texttt{get\_size}

\subsection{move\_to\_position}
\begin{figure}[h]
    \centering
    \begin{subfigure}[b]{0.33\textwidth}
        \includegraphics[width=\linewidth]{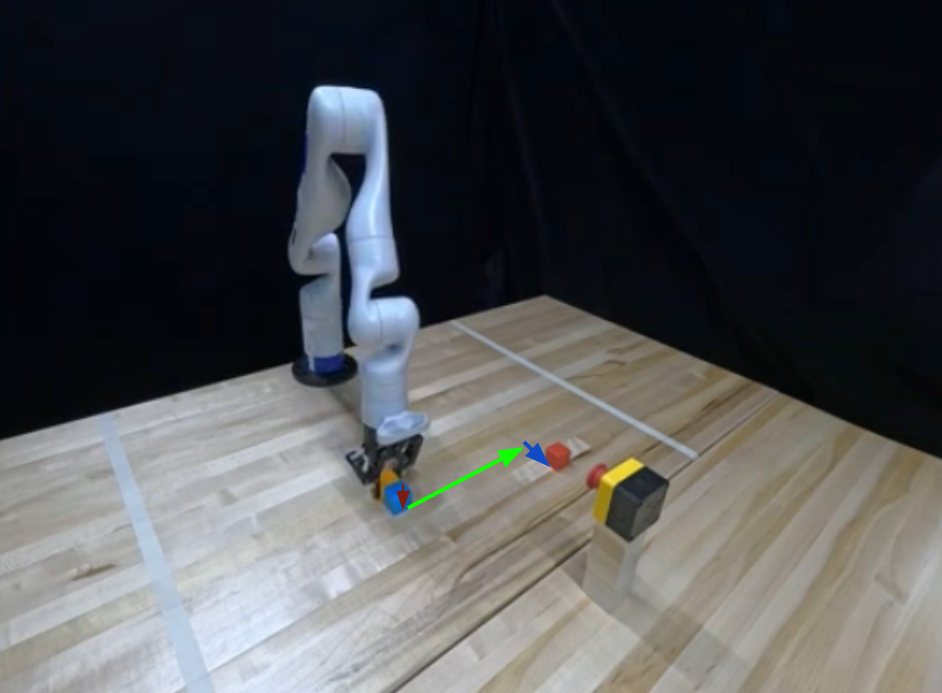}
        \caption{Button-Pressing}
    \end{subfigure}
    \begin{subfigure}[b]{0.325\textwidth}
        \includegraphics[width=\linewidth]{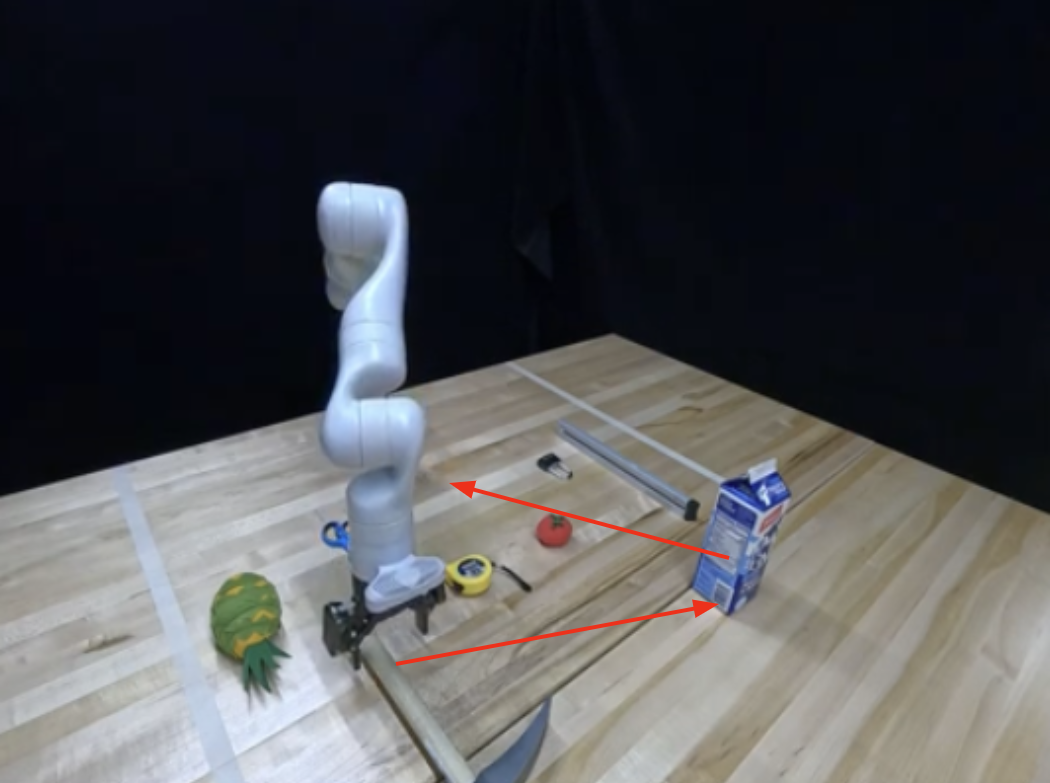}
        \caption{Milk-Reaching}
    \end{subfigure}
    \begin{subfigure}[b]{0.32\textwidth}
        \includegraphics[width=\linewidth]{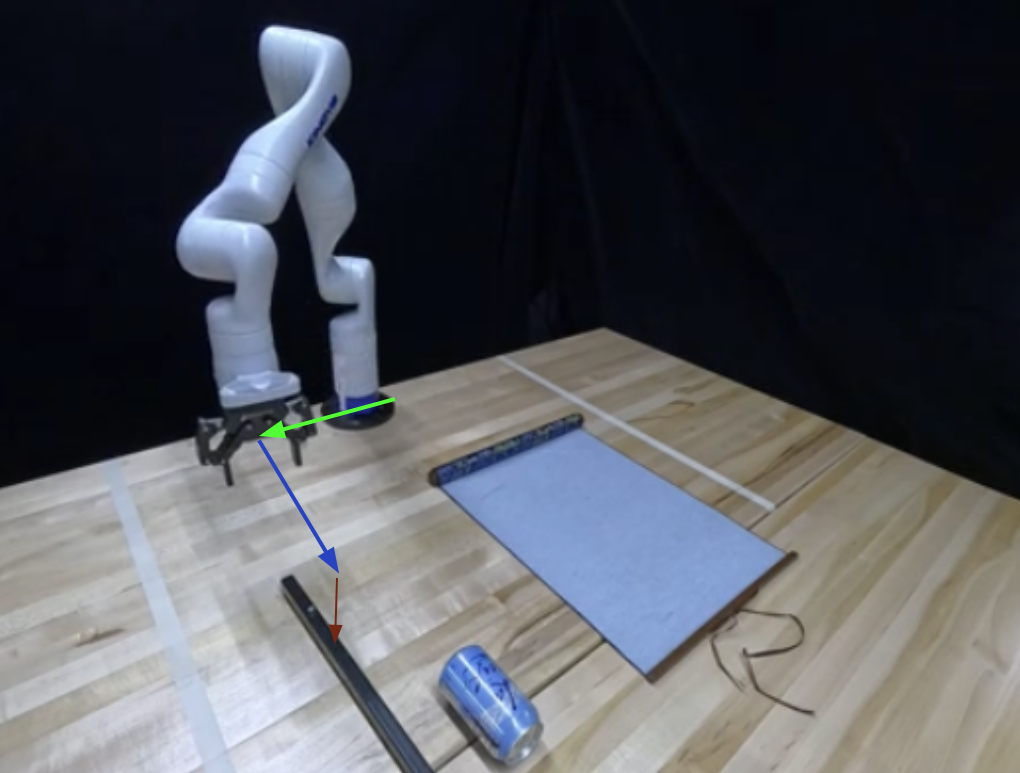}
        \caption{Can-Grasping}
    \end{subfigure}
    \caption{The figure demonstrates the scripted execution order for an arm motion planner. The red arrows in (b) show the Euclidean distance vector motion. (a) and (c) shows the scripted moving orders on each axis.}
    \label{fig:arm_motion_planner}
\end{figure}

Upon invoking the \texttt{move\_to\_position} skill, the built-in Kinova high-level planner would generate waypoints along the Euclidean distance direction between the current tool pose and target position. However, there are some constraints introduced by certain object scenes. The detailed motion planning paths are shown in Fig. \ref{fig:arm_motion_planner} and described as follows: 

\begin{enumerate}
    \item \textbf{Milk-Reaching:} Due to the geometric features of the object \textit{hammer}, which its center does not represent the grasping point of the object, we added an object-specific offset in both x and y axes to the motion planner when grasping the hammer. All the other motion behaviors are generated by \method and directly executed by the Kinova high-level motion planner. 
    
    \item \textbf{Can-Grasping:} Under the object settings, we have pre-scripted a collision-free path given the target position. Instead of moving along the Euclidean distance vector, we assume the scene is in grid world settings where the arm can only move in one direction once. The motion of approaching target objects starts with Y, followed by X, and then Z. 

    \item \textbf{Button-Pressing:} For the magnetic cube geometries, only the flat surface can be attached firmly. To resolve the instability, we assume the Button-Pressing scene is in grid world settings where the arm agent can only move in one direction once. The motion of approaching target objects starts with Z, followed by Y, and then X. 
\end{enumerate}

\subsection{get\_position}
When invoking the \texttt{get\_position} function, we employed the OWL-ViT methodology as detailed in the reference \cite{minderer2022simple}. This approach allowed us to derive 2D bounding boxes encompassing the objects within the scene. This was achieved by capturing a slightly slanted top-down perspective of the environment. Following this, we conducted a conversion of the bounded image's coordinates from 2D to both world coordinates and robot coordinates. Subsequently, we fused the depth information from stereo input to each identified object with the transformed 2D bounding box represented in robot coordinates. As a result, we were able to calculate the comprehensive 3D positions of the objects within the scene. add detection picture. Fig.~\ref{fig:kinova_owl_vit_results} presents an example of various object positions as detected by the OWL-ViT detector.

\subsection{get\_size}
\begin{figure}[htb]
\centering
\includegraphics[width=0.95\linewidth]{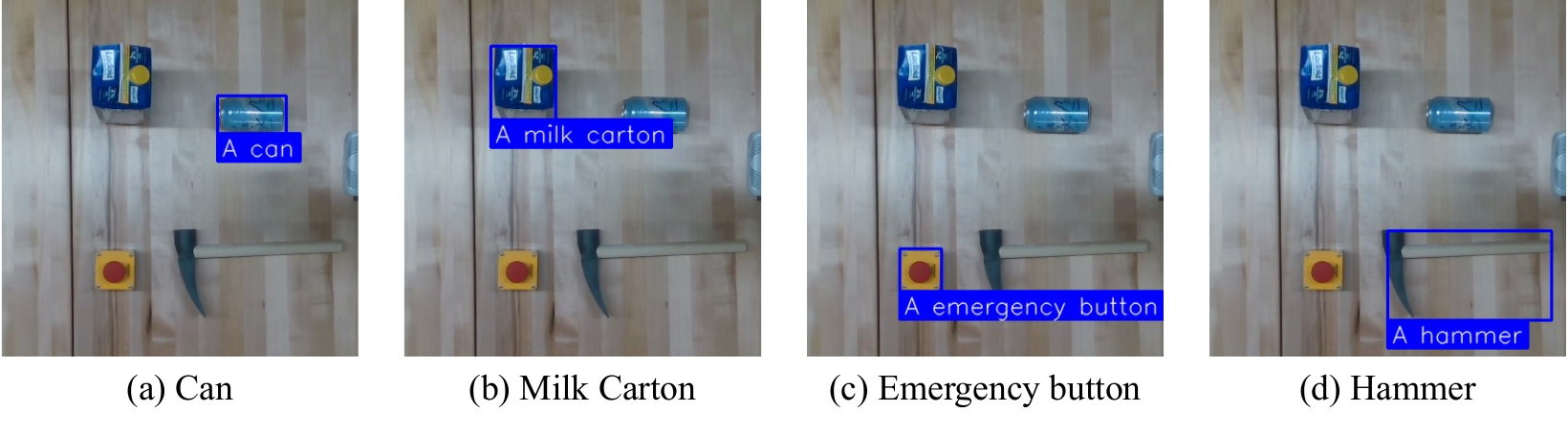}
\caption{This image demonstrates the detection capabilities of the OWL-ViT detector. The detector successfully identifies various objects along with their respective positions and bounding boxes on the table.}

\label{fig:kinova_owl_vit_results}
\end{figure}

While this function is invoked, the size of objects can also be captured using methods as described in function \texttt{get\_position}. The output is in three dimensions, which include the width, length, and height of the objects.
Fig.~\ref{fig:kinova_owl_vit_results} also presents the bounding box of various object positions.

\subsection{open \& close gripper}
This would connect with Kinova API on closing and opening the gripper. While closing the gripper, the gripper finger distance does not need explicit scripts, where the Kinova built-in gripper sensors would automatically grasp the object under pre-set pressure.

\section{Prompts}
\label{sec:appendix_prompts}

\subsection{Robotic Arms}
Prompt for \textit{Analyzer}: \href{https://raw.githubusercontent.com/Creative-RoboTool/Creative-RoboTool.github.io/main/src/texts/arm_prompt_analyzer.txt}{Link}.

Prompt for \textit{Planner}:
\href{https://raw.githubusercontent.com/Creative-RoboTool/Creative-RoboTool.github.io/main/src/texts/arm_prompt_planner.txt}{Link}.

Prompt for \textit{Calculator}:
\href{https://raw.githubusercontent.com/Creative-RoboTool/Creative-RoboTool.github.io/main/src/texts/arm_prompt_calculator.txt}{Link}.

Prompt for \textit{Coder}:
\href{https://raw.githubusercontent.com/Creative-RoboTool/Creative-RoboTool.github.io/main/src/texts/arm_prompt_coder.txt}{Link}.

\subsection{Quadrupedal Robots}
Prompt for \textit{Analyzer}:
\href{https://raw.githubusercontent.com/Creative-RoboTool/Creative-RoboTool.github.io/main/src/texts/legRobot_prompt_analyzer.txt}{Link}.

Prompt for \textit{Planner}:
\href{https://raw.githubusercontent.com/Creative-RoboTool/Creative-RoboTool.github.io/main/src/texts/legRobot_prompt_planner.txt}{Link}.

Prompt for \textit{Calculator}:
\href{https://raw.githubusercontent.com/Creative-RoboTool/Creative-RoboTool.github.io/main/src/texts/legRobot_prompt_calculator.txt}{Link}.

Prompt for \textit{Coder}:
\href{https://raw.githubusercontent.com/Creative-RoboTool/Creative-RoboTool.github.io/main/src/texts/legRobot_prompt_coder.txt}{Link}.

\section{Task Descriptions}
\label{sec:appendix_task_description}

Descriptions for \textit{Milk-Reaching}:
\href{https://raw.githubusercontent.com/Creative-RoboTool/Creative-RoboTool.github.io/main/src/texts/arm_description_hammer.txt}{Link}.

Descriptions for \textit{Can-Grasping}:
\href{https://raw.githubusercontent.com/Creative-RoboTool/Creative-RoboTool.github.io/main/src/texts/arm_description_scroll.txt}{Link}.

Descriptions for \textit{Button-Pressing}:
\href{https://raw.githubusercontent.com/Creative-RoboTool/Creative-RoboTool.github.io/main/src/texts/arm_description_magnet.txt}{Link}.

Descriptions for \textit{Sofa-Traversing}:
\href{https://raw.githubusercontent.com/Creative-RoboTool/Creative-RoboTool.github.io/main/src/texts/legRobot_description_surfboard.txt}{Link}.

Descriptions for \textit{Sofa-Climbing}:
\href{https://raw.githubusercontent.com/Creative-RoboTool/Creative-RoboTool.github.io/main/src/texts/legRobot_description_box.txt}{Link}.

Descriptions for \textit{Cube-Lifting}:
\href{https://raw.githubusercontent.com/Creative-RoboTool/Creative-RoboTool.github.io/main/src/texts/legRobot_description_lever.txt}{Link}.

\end{document}